\documentclass[journal]{IEEEtran}

\usepackage{times}
\usepackage{epsfig}
\usepackage{graphicx}
\usepackage{amsmath}
\usepackage{amssymb}
\usepackage{makecell}
\usepackage{multirow}
\usepackage{subcaption}
\usepackage{bbm}
\usepackage{verbatim}
\usepackage{cite}
\usepackage{color}
\usepackage{marvosym}
\usepackage{hyperref}
\hypersetup{
    colorlinks=true,
}
\ifCLASSINFOpdf
\else
\fi
\begin{document}
%
\title{Dual Spoof Disentanglement Generation for Face Anti-spoofing with Depth Uncertainty Learning}
%
%
%

\author{Hangtong~Wu$^*$,
        Dan~Zeng$^*$,~\IEEEmembership{Member,~IEEE,}
        Yibo~Hu\textsuperscript{\Letter},
        Hailin~Shi,~\IEEEmembership{Member,~IEEE,}
        and~Tao~Mei,~\IEEEmembership{Fellow,~IEEE}
\thanks{Hangtong Wu and Dan Zeng are with Shanghai University, shanghai 200444, China. E-mail: \{harton, dzeng\}@shu.edu.cn.}
\thanks{Yibo Hu, Hailin Shi and Tao Mei are with JD AI Research, Beijing 100020, China. E-mail: huyibo871079699@gmail.com, shihailin@jd.com, tmei@live.com}
\thanks{$^*$Equal contribution.}
\thanks{Corresponding author: Yibo Hu.}
}

%
%

\markboth{Journal of \LaTeX\ Class Files,~Vol.~14, No.~8, August~2015}%
{Shell \MakeLowercase{\textit{et al.}}: Bare Demo of IEEEtran.cls for IEEE Journals}
%



\maketitle

\begin{abstract}
Face anti-spoofing (FAS) plays a vital role in preventing face recognition systems from presentation attacks. Existing face anti-spoofing datasets lack diversity due to the insufficient identity and insignificant variance, which limits the generalization ability of FAS model. In this paper, we propose Dual Spoof Disentanglement Generation (DSDG) framework to tackle this challenge by ``anti-spoofing via generation”. Depending on the interpretable factorized latent disentanglement in Variational Autoencoder (VAE), DSDG learns a joint distribution of the identity representation and the spoofing pattern representation in the latent space. Then, large-scale paired live and spoofing images can be generated from random noise to boost the diversity of the training set. However, some generated face images are partially distorted due to the inherent defect of VAE. Such noisy samples are hard to predict precise depth values, thus may obstruct the widely-used depth supervised optimization. To tackle this issue, we further introduce a lightweight Depth Uncertainty Module (DUM), which alleviates the adverse effects of noisy samples by depth uncertainty learning. DUM is developed without extra-dependency, thus can be flexibly integrated with any depth supervised network for face anti-spoofing. We evaluate the effectiveness of the proposed method on five popular benchmarks and achieve state-of-the-art results under both intra- and inter- test settings. The codes are available at \href{https://github.com/JDAI-CV/FaceX-Zoo/tree/main/addition_module/DSDG}{\textit{https://github.com/JDAI-CV/FaceX-Zoo/tree/main/addition\_module/DSDG}}.
\end{abstract}

\begin{IEEEkeywords}
Face Anti-Spoofing, Dual Spoof Disentanglement Generation, Depth Uncertainty Learning.
\end{IEEEkeywords}

%
\IEEEpeerreviewmaketitle

\section{Introduction}
%
%
%
%
\IEEEPARstart{F}{ace} recognition system has made remarkable progress and been widely applied in various scenarios~\cite{Low2019MultiFoldGP, SepasMoghaddam2020ADS, Cevikalp2020FaceRB, Yang2021OrthogonalityLL, Wang2021FaceXZooAP, Wu2020LearningAE}. However, it is vulnerable to physical presentation attacks~\cite{Liu2020CrossethnicityFA}, such as print attack, replay attack or 3D-mask attack \textit{etc}. To make matters worse, these spoof mediums can be easily manufactured in practice. Thus, both academia and industry have paid extensive attention to developing face anti-spoofing (FAS) technology for securing the face recognition system~\cite{Yu2021DeepLF}.
\par
\begin{figure}[htp]
\centering
\includegraphics[scale=0.52]{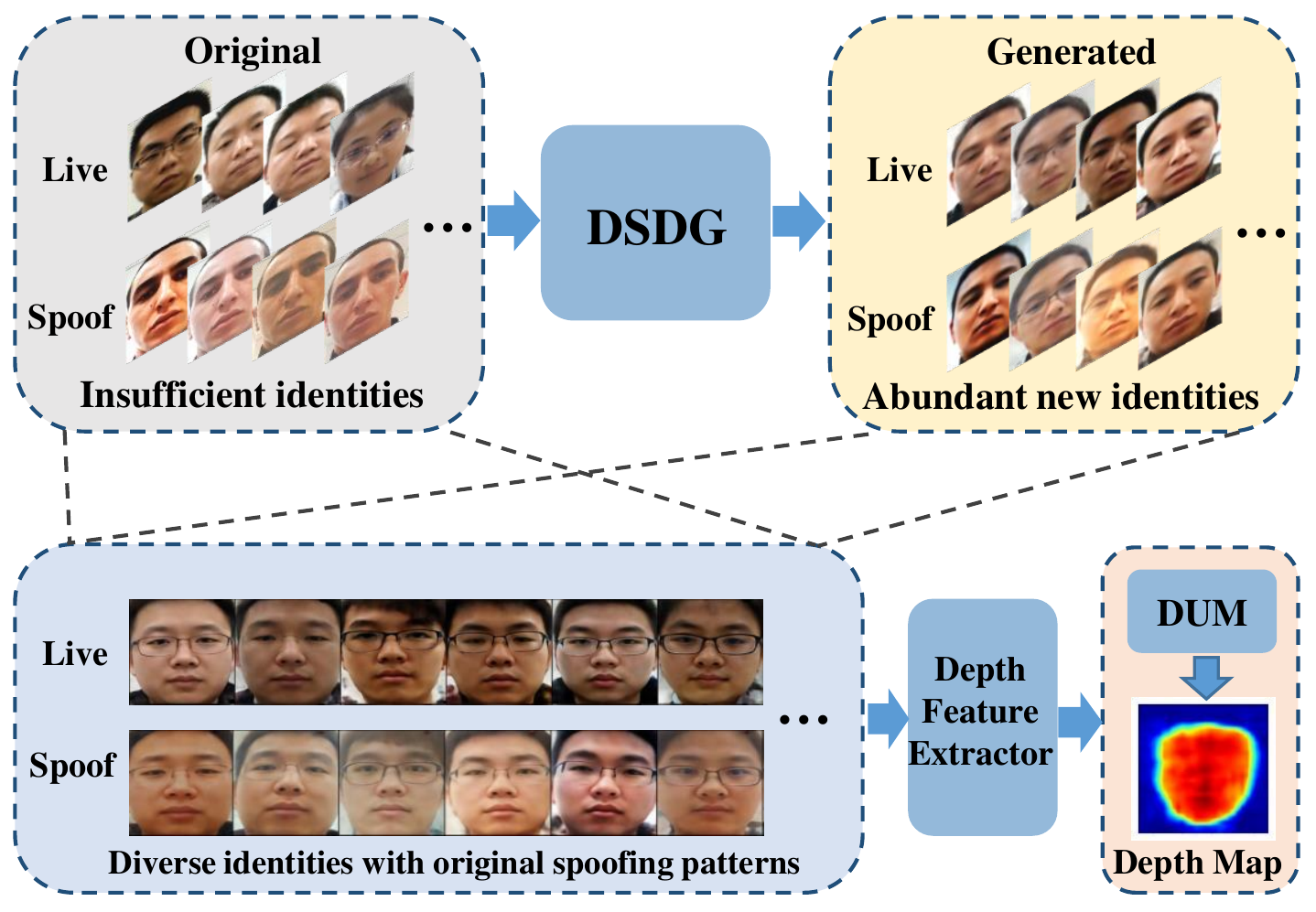}
\caption{Dual Spoof Disentanglement Generation framework can generate large-scale paired live and spoofing images to boost the diversity of the training set. Then, the extended training set is directly utilized to enhance the training of anti-spoofing network with a Depth Uncertainty Module.}
\label{ideapic}
\vspace{-1.5em}
\end{figure}
Previous works on face anti-spoofing mainly focus on how to obtain discriminative information between the live and spoofing faces. The texture-based methods~\cite{tiago2012lbp,jukka2011face,jukka2013context,yang2013face,keyu2016secure,Boulkenafet2016FaceSD} leverage textural features to distinguish spoofing faces. Besides, several works~\cite{Li2016GeneralizedFA, Liu2018learning, Lin2019FaceLD, Kim2019BASNEF} are designed based on liveness cues (\textit{e.g.} rPPG, reflection) for dynamic discrimination. Recent works~\cite{Yu2020SearchingCD, Atoum2017FaceAU, Liu2018learning, Li2016AnOF, Yu2021DualCrossCD,Shao2019JointDL,Li2018LearningGD, George2021LearningOC} employ convolution neural networks (CNNs) to extract discriminative features, which have made great progress in face anti-spoofing. However, most existing anti-spoofing datasets are insufficient in subject numbers and variances. For instance, the commonly used OULU-NPU~\cite{Boulkenafet2017OULUNPUAM} and SiW~\cite{Liu2018learning} only have 20 and 90 subjects in the training set, respectively. As a consequence, models may easily suffer from over-fitting issue, thus lack the generalization ability to the unseen target or the unknown attack. 
\par
To overcome the above issue, Yang \textit{et al.}~\cite{Yang2019FaceAM} propose a data collection solution along with a data synthesis technique to obtain a large amount of training data, which well reflects the real-world scenarios. However, this solution requires to collect data from external data sources, and the data synthesis technique is time-consuming and inefficient. Liu \textit{et al.}~\cite{Liu2020OnDS} design an adversarial framework to disentangle the spoof trace from the input spoofing face. The spoof trace is deformed towards the live face in the original dataset to synthesize new spoofing faces as the external data for training the generator in a supervised fashion. However, their method performs image-to-image translation and can only synthesize new spoof images with the same identity attributes, thus the data still lacks inter-class diversity. Considering the above limitations, we try to solve this issue from the perspective of sample generation. The overall solution is presented in Fig.~\ref{ideapic}. We first train a generator with original dataset to obtain new face images contain abundant new identities and original spoofing patterns. Then, both the original and generated face images are utilized to enhance the training of FAS network in a proper way. Specifically, our method can generate large-scale live and spoofing images with diverse variances from random noise, and the images are not limited in the same identity attributes. Moreover, the training process of the generator does not require external data sources.
\par
In this paper, we propose a novel Dual Spoof Disentanglement Generation (DSDG) framework to enlarge the intra- and inter-class diversity without external data. Inspired by the promising capacity of Variational Autoencoder (VAE)~\cite{Kingma2014AutoEncodingVB} in the interpretable latent disentanglement, we first adopt an VAE-like architecture to learn the joint distribution of the identity information and the spoofing patterns in the latent space. Then, the trained decoder can be utilized to generate large-scale paired live and spoofing images with the noise sampled from standard Gaussian distribution as the input to boost the diversity of the training set. It is worth mentioning that the generated face images contain diverse identities and variances as well as reverse original spoofing patterns. Different from~\cite{Liu2020OnDS}, our method performs generation from noise, \textit{i.e.} noise-to-image generation, and can generate both live and spoofing images with new identities. Superior to~\cite{Yang2019FaceAM}, our methods only rely on the original data source, but not external data source. However, we observe that a small portion of generated face images are partially distorted due to the inherent limitation of VAE that some generated samples are blurry and low quality. Such noisy samples are difficult to predict precise depth values, thus may obstruct the widely used depth supervised training optimization. To solve this issue, we propose a Depth Uncertainty Module (DUM) to estimate the confidence of the depth prediction. During training, the predicted depth map is not deterministic any more, but sampled from a dynamic depth representation, which is formulated as a Gaussian distribution with learned mean and variance, and is related to the depth uncertainty of the original input face. In the inference stage, only the deterministic mean values are employed as the final depth map to distinguish the live faces from the spoofing faces. It is worth mentioning that DUM can be directly integrated with any depth supervised FAS networks, and we find in experiments that DUM can also improve the training with real data. Extensive experiments on multiple datasets and protocols are conducted to demonstrate the effectiveness of our method.
In summary, our contributions lie in three folds:
\begin{itemize}
\item We propose a Dual Spoof Disentanglement Generation (DSDG) framework to learn a joint distribution of the identity representation and the spoofing patterns in the latent space. Then, a large-scale new paired live and spoofing image set is generated from random noises to boost the diversity of the training set.
\item To alleviate the adverse effects of some generated noisy samples in the training of FAS models, we introduce a Depth Uncertainty Module (DUM) to estimate the reliability of the predicted depth map by depth uncertainty learning. We further observe that DUM can also improve the training when only original data involved.
\item Extensive experiments on five popular FAS benchmarks demonstrate that our method achieves remarkable improvements over state-of-the-art methods on both intra- and inter- test settings. 
\end{itemize}
\begin{figure*}[htp]
\centering
\includegraphics[scale=0.62]{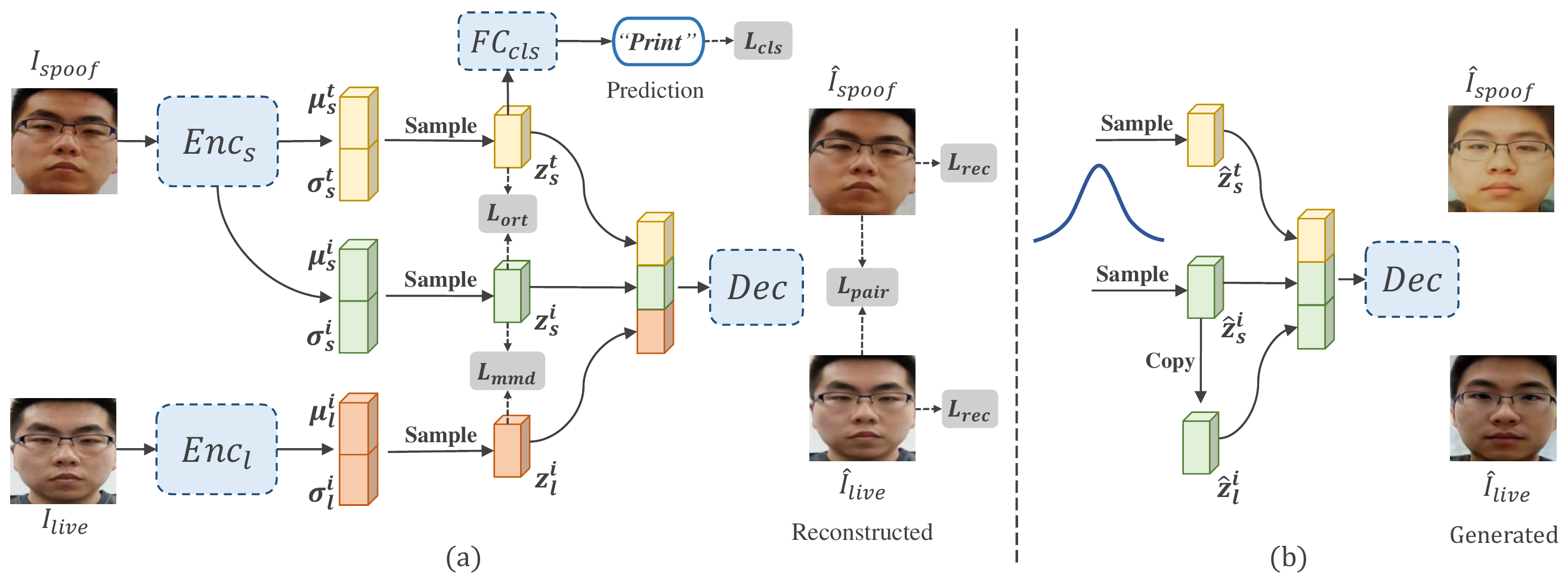}
\caption{The overview of DSDG: (a) the training process and (b) the generation process.
First, the Dual Spoof Disentanglement VAE is trained towards a joint distribution of the spoofing pattern representation and the identity representation. Then, the trained decoder in (a) is utilized in (b) to generate large number of paired images from the randomly sampled standard Gaussian noise.}
\label{DSDGpic}
\end{figure*}
\section{Related Worked}
\textbf{Face Anti-Spoofing.}
 The early traditional face anti-spoofing methods mainly rely on hand-crafted features to distinguish the live faces from the spoofing faces, such as LBP~\cite{tiago2012lbp, tiago2013can, jukka2011face}, HoG~\cite{jukka2013context, yang2013face}, SIFT~\cite{keyu2016secure} and SURF~\cite{zine2017face}. However, these methods are sensitive to the variation of illumination, pose, \textit{etc}. Taking the advantage of CNN's strong representation ability, many CNN-based methods are proposed recently. Similar to the early works, ~\cite{keyu2016secure, Li2016AnOF, Jourabloo2018FaceDA, George2019DeepPB, Shao2019JointDL, George2020BiometricFP, George2021LearningOC} extract the spatial texture with CNN models from the single frame. Others attempt to utilize CNN to obtain discriminative features. For instance, Zhu \textit{et al.}~\cite{Zhu2021DetectionOS} propose a general Contour Enhanced Mask R-CNN model to detect the spoofing medium contours from the image. Chen \textit{et al.}~\cite{Chen2021CameraIF} consider the inherent variance from acquisition cameras at the feature level for generalized FAS, and design a two CNN branches network to learn the camera invariant spoofing features and recompose the low/high-frequency components, respectively. Besides,~\cite{Lin2018LiveFV, Yang2019FaceAM, Wang2020DeepSG, Li2018LearningGD} combine the spatial and temporal textures from the frame sequence to learn more distinguishing features between the live and the spoofing faces. Specifically, Zheng \textit{et al.}~\cite{Zheng2021AttentionBasedSM} present a two-stream network spatial-temporal network with a scale-level attention module, which joints the depth and multi-scale information to extract the essential discriminative features. Meanwhile, some auxiliary supervised signals are adopted to enhance the model's robustness and generalization, such as rPPG~\cite{Li2016GeneralizedFA, Liu2018learning, Lin2019FaceLD}, depth map~\cite{Yu2020SearchingCD, Atoum2017FaceAU, Liu2018learning, Yu2020FaceAW, Wu2021SingleShotFA},~\cite{Yu2020NASFASSC, Liu2021CASIASURFCA} and reflection~\cite{Kim2019BASNEF}. Recently, some novel methods are introduced to FAS, for instance, Quan \textit{et al.}~\cite{Quan2021ProgressiveTL} propose a semi-supervised learning based adaptive transfer mechanism to obtain more reliable pseudo-labeled data to learn the FAS model. Deb \textit{et al.}~\cite{Deb2021LookLI} introduce a Regional Fully Convolutional Network to learn local cues in a self-supervised manner. Cai \textit{et al.}~\cite{Cai2021DRLFASAN} utilize deep reinforcement learning to extract discriminative local features by modeling the behavior of exploring face-spoofing-related information from image sub-patches. Yu \textit{et al.}~\cite{Yu2021RevisitingPS} present a pyramid supervision to provide richer multi-scale spatial context for fine-grained supervision, which is able to plug into existed pixel-wise supervision framework. Zhang \textit{et al.}~\cite{Zhang2021StructureDA} decompose images into patches to construct a non-structural input and recombine patches from different subdomains or classes. Besides, ~\cite{Shao2019MultiAdversarialDD, Jia2020SingleSideDG, Chen2021GeneralizableRL, Wang2021SelfDomainAF, Liu2021DualRD,Yang2015PersonSpecificFA,Li2018UnsupervisedDA,Wang2021UnsupervisedAD},~\cite{Liu2021AdaptiveNR} introduce domain generation into the FAS task to alleviate the poor generalizability to unseen domains. Qin \textit{et al.}~\cite{Qin2020LearningMM, Qin2021MetateacherFF} utilize meta-teacher to supervise the presentation attack detector to learning rich spoofing cues.
 \par
Recently, researchers pay more attention to solving face anti-spoofing with the generative model. Jourabloo \textit{et al.}~\cite{Jourabloo2018FaceDA} decompose a spoofing face image into a live face and a spoof noise, then utilize the spoof noise for classification. Liu \textit{et al.}~\cite{Liu2021FaceAV} present a GAN-like image-to-image framework to translate the face image from RGB to NIR modality. Then, the discriminative feature of RGB modality can be obtained with the assistance of NIR modality to further promote the generalization ability  of FAS model. Inspired by the disentangled representation, Zhang \textit{et al.}~\cite{Zhang2020FaceAV} adopt GAN-like discriminator to separate liveness features from the latent representations of the input faces for further classification. Liu \textit{et al.}~\cite{Liu2020OnDS} disentangle the spoof traces to reconstruct the live counterpart from the spoof faces and synthesize new spoof samples from the live ones. The synthesized spoof samples are further employed to train the generator. Finally, the intensity of spoof traces are used for prediction. 
\par
\textbf{Generative Model.}
Variational autoencoders (VAEs)~\cite{Kingma2014AutoEncodingVB} and generative adversarial networks (GANs)~\cite{Goodfellow2014GenerativeAN} are the most basic generative models. VAEs have promising capacity in latent representation, which consist of a generative network (Decoder) $p_{\theta}{(x\mid z)}$ and an inference network (Encoder) $q_{\phi}{(z\mid x)}$. The decoder generates the visible variable $x$ given the latent variable $z$, and the encoder maps the visible variable $x$ to the latent variable $z$ which approximates $p(z)$. Differently, GANs employ a generator and a discriminator to implement a min-max mechanism. On one hand, the generator generates images to confuse the discriminator. On the other hand, the discriminator tends to distinguish between generated images and real images. Recently, several works have introduced the "X via generation" manner to facial sub-tasks. For instance, ``recognition via generation"~\cite{Zhao2018TowardsPI, Zhao2019LookAE, Fu2019DualVG}, ``parsing via generation"~\cite{Li2020DualStructureDV} and others~\cite{Hu2018PoseGuidedPF, Cao2018LearningAH, Zhao2017DualAgentGF, Cao20193DAD, Fu2021HighFidelityFM, Li2019M2FPAAM}. In this paper, we consider the interpretable factorized latent disentanglement of VAEs and explore ``anti-spoofing via generation".
\par
\textbf{Uncertainty in Deep Learning.}
In recent years, lots of works begin to discuss what role uncertainty plays in deep learning from the theoretical perspective~\cite{Blundell2015WeightUI, Gal2016DropoutAA, Kendall2017WhatUD}. Meanwhile, uncertainty learning has been widely used in computer vision tasks to improve the model robustness and interpretability, such as face analysis~\cite{Khan2019StrikingTR, Shi2019ProbabilisticFE, Chang2020DataUL, She2021DiveIA}, semantic segmentation~\cite{Isobe2017DeepCE, Kendall2017BayesianSM} and object detection~\cite{Choi2019GaussianYA, Kraus2019UncertaintyEI}. However, most methods focus on capturing the noise of the parameters by studying the model uncertainty. In contrast, Chang~\textit{et al.}~\cite{Chang2020DataUL} apply data uncertainty learning to estimate the noise inherent in the training data. Specifically, they map each input image to a Gaussian Distribution, and simultaneously learn the identity feature and the feature uncertainty. Inspired by~\cite{Chang2020DataUL}, we treat the minor partial distortion during data generation as a kind of noise, which is hard to predict precise depth value, and introduce the depth uncertainty to capture such noise. To the best of our knowledge, this is the first to utilize uncertainty learning in face anti-spoofing tasks.
\par
\section{Methodology}
Previous approaches rarely consider face anti-spoofing from the perspective of data, while the existing FAS datasets usually lack the visual diversity due to the limited identities and insignificant variance. The OULU-NPU~\cite{Boulkenafet2017OULUNPUAM} and SiW~\cite{Liu2018learning} only contain 20 and 90 identities in the training set, respectively. To mitigate the above issue and increase the intra- and inter-class diversity, we propose a Dual Spoof Disentanglement Generation (DSDG) framework to generate large-scale paired live and spoofing images without external data acquisition. In addition, we also investigate the limitation of the generated images, and develop a Depth Uncertainty Learning (DUL) framework to make better use of the generated images. In summary, our method aims to solve the following two problems: (1) \textit{how to generate diverse face data for anti-spoofing task without external data acquisition}, and (2) \textit{how to properly utilize the generated data to promote the training of face anti-spoofing models.} 
Correspondingly, we first introduce DSDG in Sec.~\ref{DSDG}, then describe DUL for face anti-spoofing in Sec.~\ref{DUM}. 
\par
\subsection{Dual Spoof Disentanglement Generation}
\label{DSDG}
Given the pairs of live and spoofing images from the limited identities, our goal is to train a generator which can generate diverse large-scale paired data from random noise. To achieve this goal, we propose a Dual Spoof Disentanglement VAE, which accomplishes two objectives: 1) preserving the spoofing-specific patterns in the generated spoofing images, and 2) guaranteeing the identity consistency of the generated paired images. The details of Dual Spoof Disentanglement VAE and the data generation process are depicted as follows.
\smallskip
\subsubsection{Dual Spoof Disentanglement VAE}
\label{dsda}
The structure of Dual Spoof Disentanglement VAE is illustrated in Fig.~\ref{DSDGpic}(a). It consists of two encoder networks, a decoder network and a spoof disentanglement branch. The encoder $Enc_{l}$ and $Enc_{s}$ are adopted to maps the paired images to the corresponding distributions. Specifically, $Enc_{l}$ maps the live images to the identity distribution $z^{i}_{l}$. $Enc_{s}$ maps the spoofing images to the spoofing pattern distribution $z^{t}_{s}$ and the identity distribution $z^{i}_{s}$ in the latent space. The processes can be formulated as:
\begin{equation}
\begin{split}
&{z^{i}_{l}}=q_{\phi_{l}}\left(z^{i}_{l} \mid I_{live}\right),
\end{split}
\end{equation}
\begin{equation}
\begin{split}
&{z^{t}_{s}}=q_{\phi_{s}}\left(z^{t}_{s} \mid I_{spoof}\right),
\end{split}
\end{equation}
\begin{equation}
\begin{split}
&{z^{i}_{s}}=q_{\phi_{s}}\left(z^{i}_{s} \mid I_{spoof}\right),
\label{map process}
\end{split}
\end{equation}
where $q_{\phi}(\cdot)$ represents the posterior distribution. $\phi_{l}$ and $\phi_{s}$ denote the parameters of $Enc_{l}$ and $Enc_{s}$, respectively.
\par
To instantiate these processes, we follow the reparameterization routine in~\cite{Kingma2014AutoEncodingVB}. Taking the encoder $Enc_{l}$ as an example, instead of directly obtaining the identity distribution $z^{i}_{l}$, $Enc_{l}$ outputs the mean $\mu^{i}_{l}$ and the standard deviation $\sigma^{i}_{l}$ of $z^{i}_{l}$. Subsequently, the identity distribution can be obtained by: $z^{i}_{l}=\mu^{i}_{l}+\epsilon\sigma^{i}_{l}$, where $\epsilon$ is a random noise sampled from a standard Gaussian distribution. Similar processes are also conducted on $z^{t}_{s}$ and $z^{i}_{s}$, respectively. After obtaining the $z^{i}_{l}$, $z^{t}_{s}$ and $z^{i}_{s}$, we concatenate them to a feature vector and feed it to the decoder $Dec$ to generate the reconstructed paired images $ \hat{I}_{live} $ and $ \hat{I}_{spoof} $.
\par
\textbf{Spoof Disentanglement.}
Generally, the input spoofing images contain multiple spoof types (\textit{e.g.} print and replay \textit{etc.}). To ensure the generated spoofing image $ \hat{I}_{spoof} $ preserves the spoof information, it is crucial to disentangle the spoofing image into the spoofing pattern representation and the identity representation. Therefore, the encoder $Enc_{s}$ is designed to map the spoofing image $ I_{spoof} $ into two distributions: $ z^{t}_{s}$ and $ z^{i}_{s} $, \textit{i.e.} the spoofing pattern representation and the identity representation in the latent space, respectively. To better learn the spoofing pattern representation $ z^{t}_{s}$, we adopt a classifier to predict the spoofing type by minimizing the following CrossEntropy loss:
\begin{equation}
\begin{split}
\mathcal{L}_{\mathrm{cls}}=\operatorname{CrossEntropy}\left(fc(z^{t}_{s}),\; y\right),
\label{cls loss}
\end{split}
\end{equation}
where $fc(\cdot)$ represents a fully connected layer, and y is the label of the spoofing type.
\par
In addition, an angular orthogonal constraint is also adopted between $z^{t}_{s}$ and $z^{i}_{s}$ to guarantee the $Enc_{s}$ effectively disentangle the $ I_{spoof} $ into the spoofing pattern representation and the identity representation. The angular orthogonal loss is formulated as:
\begin{equation}
\begin{split}
\mathcal{L}_{\text {ort}}=\left|\left\langle\frac{z_{s}^{t}}{\left\|z_{s}^{t}\right\|_{2}}, \frac{z_{s}^{i}}{\left\|z_{s}^{i}\right\|_{2}}\right\rangle\right|,
\label{ort2}
\end{split}
\end{equation}
where $\langle\cdot, \cdot\rangle$ denotes inner product. By minimizing $\mathcal{L}_{\text {ort}}$, $z^{t}_{s}$ and $z^{i}_{s}$ are constrained to be orthogonal, forcing the spoofing pattern representation and identity representation to be disentangled.
\par
\textbf{Distribution Learning.}
We employ a VAE-like network to learn the joint distribution of the spoofing pattern representation and the identity representation. The posterior distributions $q_{\phi_{s}}\left(z^{t}_{s} \mid I_{spoof}\right)$, $q_{\phi_{s}}\left(z^{i}_{s} \mid I_{spoof}\right)$ and $q_{\phi_{l}}\left(z^{i}_{l} \mid I_{live}\right)$ are constrained by the Kullback-Leibler divergence:
\par
\begin{equation}
\begin{split}
\mathcal{L}_{\mathrm{kl}}=&D_{\mathrm{KL}}\left(q_{\phi_{s}}\left(z^{t}_{s} \mid I_{spoof}\right) \| p\left(z^{t}_{s}\right)\right)\\+&D_{\mathrm{KL}}\left(q_{\phi_{s}}\left(z^{i}_{s} \mid I_{spoof}\right) \| p\left(z^{i}_{s}\right)\right)\\+&D_{\mathrm{KL}}\left(q_{\phi_{l}}\left(z^{i}_{l} \mid I_{live}\right) \| p\left(z^{i}_{l}\right)\right).
\label{kl loss}
\end{split}
\end{equation}
\indent Given the spoofing pattern representation $z^{t}_{s}$, the identity representations  $z^{i}_{s}$ and $z^{i}_{l}$, the decoder network $\textit{Dec}$ aims to reconstruct the inputs ${I_{spoof}}$ and ${I_{live}}$:
\begin{equation}
\begin{split}
\mathcal{L}_{\mathrm{rec}}=-\mathbb{E}_{q_{\phi_{s}} \cup q_{\phi_{l}}} \log p_{\theta}\left(I_{spoof}, I_{live} \mid z^{t}_{s}, z^{i}_{s}, z^{i}_{l}\right),
\label{rec loss}
\end{split}
\end{equation}
where $\theta$ denotes the parameters of the decoder network. In practice, we constrain the $L_1$ loss between the reconstructed images $\hat{I}_{spoof}$/ $\hat{I}_{live}$ and the original images $I_{spoof}$/ $I_{live}$:
\begin{equation}
\begin{split}
\mathcal{L}_{\mathrm{rec}}=\left\|\hat{I}_{spoof}-I_{spoof}\right\|_{1}+\left\|\hat{I}_{live}-I_{live}\right\|_{1}.
\label{rec loss2}
\end{split}
\end{equation}
\par
\textbf{Distribution Alignment.}
Meanwhile, we align the identity distributions of $z^{i}_{s}$ and $z^{i}_{l}$ by a Maximum Mean Discrepancy loss to preserve the identity consistency in the latent space:
\begin{equation}
\begin{split}
\mathcal{L}_{\mathrm{mmd}}=\left|\frac{1}{n} \sum_{j=1}^{n} z^{i}_{s,j} -\frac{1}{n} \sum_{k=1}^{n} z^{i}_{l,k}\right|,
\label{mmd loss}
\end{split}
\end{equation}
where $n$ denotes the dimension of $z^{i}_{s}$ and $z^{i}_{l}$.
\par
\par
\textbf{Identity Constraint.}
To further preserve the identity consistency, an identity constraint is also employed in the image space. Similar to the previous work~\cite{Fu2019DualVG, Fu2021DVGFaceDV}, we leverage a pre-trained LightCNN~\cite{Wu2018ALC} as the identity feature extractor $F_{ip}(\cdot)$ and deploy a feature $L_2$ distance loss to constrain the identity consistency of the reconstructed paired images:
\begin{equation}
\begin{split}
\mathcal{L}_{\text {pair}}=\left\|F_{ip}\left(\hat{I}_{spoof}\right)-F_{i p}\left(\hat{I}_{live}\right)\right\|_{2}^{2}.
\label{ip-pair}
\end{split}
\end{equation}
\par
\textbf{Overall Loss.}
The total objective function is a weighted sum of the above losses, defined as:
\begin{equation}
\begin{split}
\mathcal{L}=\mathcal{L}_{\text {kl }}+\mathcal{L}_{\text {rec }}+\lambda_{1} \mathcal{L}_{\text {mmd}}+\lambda_{2} \mathcal{L}_{\text {pair }}+\lambda_{3} \mathcal{L}_{\text {ort }}+\lambda_{4} \mathcal{L}_{\text {cls }},
\label{gen loss}
\end{split}
\end{equation}
where $\lambda_{1}$, $\lambda_{2}$, $\lambda_{3}$ and $\lambda_{4}$ are the trade-off parameters.
\par
\bigskip
\subsubsection{Data Generation}
\label{data gen}
\indent The procedure of generating paired live and spoofing images is shown in Fig.~\ref{DSDGpic}(b).
Similar to the reconstruction process in Sec.~\ref{dsda}, we first obtain the $\hat{z}^{t}_{s}$ and $\hat{z}^{i}_{s}$ by sampling from the standard Gaussian distribution $\mathcal{N}(0, \mathrm{I})$. In order to keep the identity consistency of the generated paired data, the identity representation $\hat{z}^{i}_{l}$ is not sampled but copied from $\hat{z}^{i}_{s}$. Then, we concatenate $\hat{z}^{t}_{s}$, $\hat{z}^{i}_{s}$ and $\hat{z}^{i}_{l}$ as a joint representation, and feed it to the trained decoder $\textit{Dec}$ to obtain the new paired live and spoofing images.
\par
It is worth mentioning that DSDG can generate large-scale paired live and spoofing images through independently repeated sampling. Some generation results of DSDG are shown in Fig.~\ref{render}, where the generated paired images contain the identities that do not exist in the real data. In addition, the spoofing images also successfully preserve the spoofing pattern from the real data. Thus, our DSDG can effectively increase the diversity of the original training data. However,it is inevitable that a portion of generated samples have some appearance distortions in the image space due to the inherent limitation of VAEs, such as blur. When directly absorbing them into the training data, such distortion in the generated image may harm the training of anti-spoofing model. To handle this issue, we propose a Depth Uncertainty Learning to reduce the negative impact of such noisy samples during training. 
\par
\begin{figure}[t]
\centering
\includegraphics[scale=0.56]{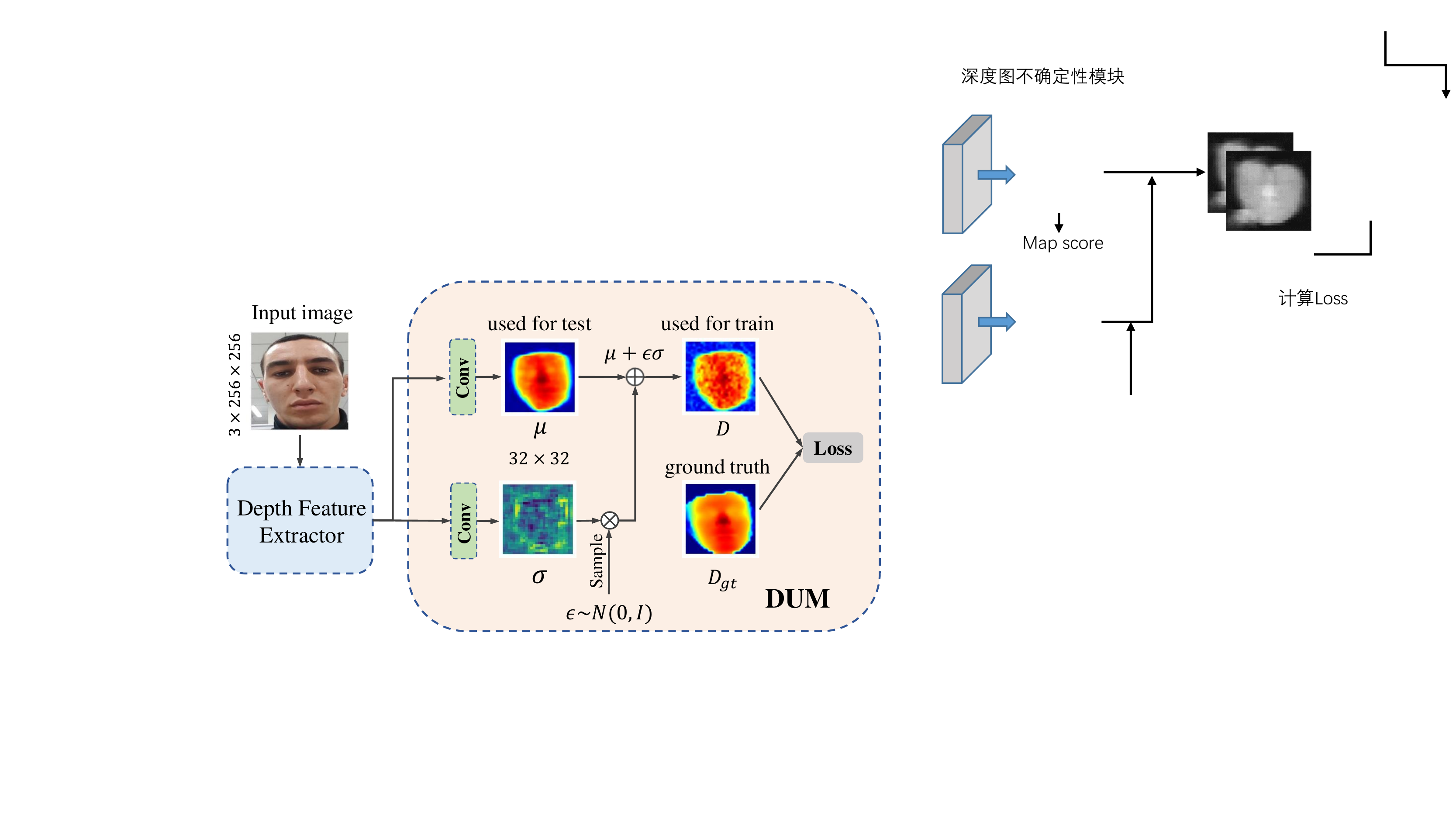}
\caption{The overview of the proposed DUM.}
\label{dumpic}
\vspace{-0.5em}
\end{figure}
\subsection{Depth Uncertainty Learning for FAS}
\label{DUM}
Facial depth map as a representative supervision information, which reflects the facial depth information with respect to different face areas, has been widely applied in face anti-spoofing methods~\cite{Liu2018learning, Yu2020SearchingCD, Wang2020DeepSG}. Typically, a depth feature extractor $F_{d}(\cdot)$ is used to learn the mapping from the input RGB image $\mathbf{X} \in \mathbb{R}^{3 \times M \times N}$ to the output depth map $\mathbf{D} \in \mathbb{R}^{m \times n}$, where M and N are $k$ times $m$ and $n$, respectively. The ground truth depth map of live face is obtained by an off-the-shelf 3D face reconstruction network, such as PRNet~\cite{Feng2018Joint3F}. For the spoofing face, we directly set the depth map to zeros following~\cite{Yu2020SearchingCD}. Each depth value $d_{i,j}$ of the output depth map $\mathbf{D}$ corresponds to the image patch $\mathbf{x}_{i,j} \in \mathbb{R}^{3 \times k \times k}$ of the input $\mathbf{X}$, where $i \in \{1,2,...,m\}$ and $j \in \{1,2,...,n\}$.
\par
\textbf{Depth Uncertainty Representation.}
As mentioned in Sec.~\ref{data gen}, a few generated images may suffer from partial facial distortion, and it is difficult to predict a precise depth value of the corresponding face area. When directly absorbing these generated images into the training set, such distortion will obstruct the training of anti-spoofing model. We introduce depth uncertainty learning to solve this issue. Specifically, for each image patch $\mathbf{x}_{i,j}$ of the input $\textbf{X}$, we no longer predict a fixed depth value $d_{i,j}$ in the training stage, but learn a depth distribution $z_{i,j}$ in the latent space, which is defined as a Gaussian distribution:
\begin{equation}
\begin{split}
p\left(\mathbf{z}_{i,j} \mid \mathbf{x}_{i,j}\right)=\mathcal{N}\left(\mathbf{z}_{i,j} ; \boldsymbol{\mu}_{i,j}, \boldsymbol{\sigma}_{i,j}^{2}\right),
\label{depth pre}
\end{split}
\end{equation}
where $\mu_{i,j}$ is the mean that denotes the expected depth value of the image patch $\mathbf{x}_{i,j}$, and $\sigma_{i,j}$ is the standard deviation that reflects the uncertainty of the predicted $\mu_{i,j}$. During training, the final depth value $d_{i,j}$ is no longer deterministic, but stochastic sampled from the depth distribution $p\left(\mathbf{z}_{i,j} \mid \mathbf{x}_{i,j}\right)$.
\par
\begin{table}
\setlength{\abovecaptionskip}{0.5cm}
\caption{Architectures of DepthNet and CDCN with DUM.}
\vspace{-1.5em}
\begin{center}
\resizebox{.45\textwidth}{!}{
\begin{tabular}{|c|l|l|l|}
\hline
\multicolumn{1}{|c|}{Layer} & \multicolumn{1}{c|}{Output} & \multicolumn{1}{c|}{DepthNet~\cite{Liu2018learning}} & \multicolumn{1}{c|}{CDCN~\cite{Yu2020SearchingCD}} \\ \hline
\multicolumn{1}{|c|}{Stem} & \multicolumn{1}{c|}{$256 \times 256$} & \multicolumn{1}{c|}{$3 \times 3$ $conv$, 64} & \multicolumn{1}{c|}{$3 \times 3$ $CDC$, 64} \\ \hline
\multirow{4}{*}{Low} & \multicolumn{1}{c|}{\multirow{4}{*}{$128 \times 128$}} & \multicolumn{1}{c|}{\multirow{4}{*}{$\begin{bmatrix} 3 \times 3 \ conv, 128 \\ 3 \times 3 \ conv, 196 \\ 3 \times 3 \ conv, 128 \\ 3 \times 3 \ max\ pool \end{bmatrix}$}} & \multicolumn{1}{c|}{\multirow{4}{*}{$\begin{bmatrix} 3 \times 3 \ CDC, 128 \\ 3 \times 3 \ CDC, 196 \\ 3 \times 3 \ CDC, 128 \\ 3 \times 3 \ max\ pool \end{bmatrix}$}} \\
 &  &  &  \\
 &  &  &  \\
 &  &  &  \\ \hline
\multirow{4}{*}{Mid} & \multicolumn{1}{c|}{\multirow{4}{*}{$64 \times 64$}} & \multirow{4}{*}{$\begin{bmatrix} 3 \times 3 \ conv, 128 \\ 3 \times 3 \ conv, 196 \\ 3 \times 3 \ conv, 128 \\ 3 \times 3 \ max\ pool \end{bmatrix}$} & \multirow{4}{*}{$\begin{bmatrix} 3 \times 3 \ CDC, 128 \\ 3 \times 3 \ CDC, 196 \\ 3 \times 3 \ CDC, 128 \\ 3 \times 3 \ max\ pool \end{bmatrix}$} \\
 &  &  &  \\
 &  &  &  \\
 &  &  &  \\ \hline
\multirow{4}{*}{High} & \multicolumn{1}{c|}{\multirow{4}{*}{$32 \times 32$}} & \multirow{4}{*}{$\begin{bmatrix} 3 \times 3 \ conv, 128 \\ 3 \times 3 \ conv, 196 \\ 3 \times 3 \ conv, 128 \\ 3 \times 3 \ max\ pool \end{bmatrix}$} & \multirow{4}{*}{$\begin{bmatrix} 3 \times 3 \ CDC, 128 \\ 3 \times 3 \ CDC, 196 \\ 3 \times 3 \ CDC, 128 \\ 3 \times 3 \ max\ pool \end{bmatrix}$} \\
 &  &  &  \\
 &  &  &  \\
 &  &  &  \\ \hline
\multicolumn{1}{|c|}{} & \multicolumn{1}{c|}{$32 \times 32$} & \multicolumn{2}{c|}{[concat(Low,\ Mid,\ High),\ 384]}  \\ \hline
\multirow{3}{*}{Head} & \multicolumn{1}{c|}{\multirow{3}{*}{$32 \times 32$}} & \multirow{3}{*}{$\begin{bmatrix} 3 \times 3 \ conv, 128 \\ 3 \times 3 \ conv, 64 \end{bmatrix}$} & \multirow{3}{*}{$\begin{bmatrix} 3 \times 3 \ CDC, 128 \\ 3 \times 3 \ CDC, 64 \end{bmatrix}$} \\
&  &  &  \\
&  &  &  \\ \hline
\multirow{4}{*}{DUM} & \multicolumn{1}{c|}{\multirow{4}{*}{$32 \times 32$}} & \multicolumn{1}{c|}{\multirow{4}{*}{\begin{tabular}[c]{@{}c@{}} $3 \times 3$ \ $conv$($\mu$), 1\\ $3 \times 3$ \ $conv$($\sigma$), 1\\ reparameterize, 1\end{tabular}}} & \multicolumn{1}{c|}{\multirow{4}{*}{\begin{tabular}[c]{@{}c@{}} $3 \times 3$ \ $conv$($\mu$), 1\\ $3 \times 3$ \ $conv$($\sigma$), 1\\ reparameterize, 1\end{tabular}}} \\
&  &  &  \\
&  &  &  \\
&  &  &  \\ \hline
\multicolumn{2}{|c|}{Params} & \multicolumn{1}{c|}{2.25M}& \multicolumn{1}{c|}{2.33M}  \\ \hline
\multicolumn{2}{|c|}{FLOPs} & \multicolumn{1}{c|}{47.43G}& \multicolumn{1}{c|}{50.96G}  \\ \hline
\end{tabular}}
\end{center}
\label{resandmb}
\vspace{-0.5em}
\end{table}
\begin{table}
\setlength{\abovecaptionskip}{0.5cm}
\caption{Architectures of modified ResNet and MobileNetV2 with DUM.}
\vspace{-1.5em}
\begin{center}
\resizebox{.45\textwidth}{!}{
\begin{tabular}{|c|l|l|l|}
\hline
\multicolumn{1}{|c|}{Layer} & \multicolumn{1}{c|}{Output} & \multicolumn{1}{c|}{ResNet~\cite{He2016DeepRL}} & \multicolumn{1}{c|}{MobileNetV2~\cite{Sandler2018MobileNetV2IR}} \\ \hline
\multicolumn{1}{|c|}{Stem} & \multicolumn{1}{c|}{$256 \times 256$} & \multicolumn{1}{c|}{$3 \times 3$ $conv$, 64} & \multicolumn{1}{c|}{$3 \times 3$ $conv$, 32} \\ \hline
\multirow{3}{*}{Stage1} & \multicolumn{1}{c|}{\multirow{3}{*}{$256 \times 256$}} & \multicolumn{1}{c|}{\multirow{3}{*}{$\begin{bmatrix} 3 \times 3, 64 \\ 3 \times 3, 64 \end{bmatrix} \times 2$}} & \multicolumn{1}{c|}{\multirow{3}{*}{$\begin{bmatrix} bottleneck, 16 \\ bottleneck, 24 \end{bmatrix}\times 1$}} \\
 &  &  &  \\
 &  &  &  \\ \hline
\multirow{3}{*}{Stage2} & \multicolumn{1}{c|}{\multirow{3}{*}{$128 \times 128$}} & \multicolumn{1}{c|}{\multirow{3}{*}{$\begin{bmatrix} 3 \times 3, 128 \\ 3 \times 3, 128 \end{bmatrix} \times 2$}} & \multicolumn{1}{c|}{\multirow{3}{*}{$\begin{bmatrix} bottleneck, 32 \end{bmatrix} \times 1$}} \\
 &  &  &  \\
 &  &  &  \\ \hline
 \multirow{3}{*}{Stage3} & \multicolumn{1}{c|}{\multirow{3}{*}{$64 \times 64$}} & \multicolumn{1}{c|}{\multirow{3}{*}{$\begin{bmatrix} 3 \times 3, 256 \\ 3 \times 3, 256 \end{bmatrix} \times 2$}} & \multicolumn{1}{c|}{\multirow{3}{*}{$\begin{bmatrix} bottleneck, 64 \\ bottleneck, 96\end{bmatrix} \times 1$}} \\
 &  &  &  \\
 &  &  &  \\ \hline
 \multirow{3}{*}{Stage4} & \multicolumn{1}{c|}{\multirow{3}{*}{$32 \times 32$}} & \multicolumn{1}{c|}{\multirow{3}{*}{$\begin{bmatrix} 3 \times 3, 512 \\ 3 \times 3, 512 \end{bmatrix} \times 2$}} & \multicolumn{1}{c|}{\multirow{3}{*}{$\begin{bmatrix} bottleneck, 160 \\ bottleneck, 320\end{bmatrix} \times 1$}} \\
 &  &  &  \\
 &  &  &  \\ \hline
\multirow{3}{*}{Head} & \multicolumn{1}{c|}{\multirow{3}{*}{$32 \times 32$}} & \multirow{3}{*}{$\begin{bmatrix} 3 \times 3 \ conv, 128 \\ 3 \times 3 \ conv, 64 \end{bmatrix}$} & \multicolumn{1}{c|}{\multirow{3}{*}{$\begin{bmatrix} 3 \times 3 \ conv, 128 \\ 3 \times 3 \ conv, 64 \end{bmatrix}$}} \\
&  &  &  \\
&  &  &  \\ \hline
\multirow{4}{*}{DUM} & \multicolumn{1}{c|}{\multirow{4}{*}{$32 \times 32$}} & \multicolumn{1}{c|}{\multirow{4}{*}{\begin{tabular}[c]{@{}c@{}}$3 \times 3$ $conv$($\mu$), 1\\ $3 \times 3$ $conv$($\sigma$), 1\\ reparameterize, 1\end{tabular}}} & \multicolumn{1}{c|}{\multirow{4}{*}{\begin{tabular}[c]{@{}c@{}}$3 \times 3$ $conv$($\mu$), 1\\ $3 \times 3$ $conv$($\sigma$), 1\\ reparameterize, 1\end{tabular}}} \\
&  &  &  \\
&  &  &  \\
&  &  &  \\ \hline
\multicolumn{2}{|c|}{Params} & \multicolumn{1}{c|}{11.47M}& \multicolumn{1}{c|}{1.18M} \\ \hline
\multicolumn{2}{|c|}{FLOPs} & \multicolumn{1}{c|}{35.94G}& \multicolumn{1}{c|}{2.51G}  \\ \hline
\end{tabular}}
\end{center}
\label{dnandcdcn}
\vspace{-1em}
\end{table}
\textbf{Depth Uncertainty Module.}
We employ a Depth Uncertainty Module (DUM) to estimate the mean $\mu_{i,j}$ and the standard deviation $\sigma_{i,j}$ simultaneously, which is presented in Fig.~\ref{dumpic}. DUM is equipped behind the depth feature extractor, including two independent convolutional layers. One is for predicting the $\mu_{i,j}$, and another is for the $\sigma_{i,j}$. Both of them are the parameters of a Gaussian distribution. However, the training is not differentiable during the gradient backpropagation if we directly sample $d_{i,j}$ from the Gaussian distribution. We follow the reparameterization routine in~\cite{Kingma2014AutoEncodingVB} to make the process learnable, and the depth value $d_{i,j}$ can be obtained as:
\begin{equation}
\begin{split}
d_{i,j}=\boldsymbol{\mu}_{i,j}+\epsilon \boldsymbol{\sigma}_{i,j}, \quad \epsilon \sim \mathcal{N}(\mathbf{0}, \mathbf{I}).
\label{depth sample}
\end{split}
\end{equation}
\par
In addition, same as~\cite{Chang2020DataUL}, we adopt a Kullback-Leibler divergence as the regularization term to constrain $\mathcal{N}\left(\boldsymbol{\mu}_{i}, \boldsymbol{\sigma}_{i,j}\right)$ to be close to a constructed distribution $\mathcal{N}\left(\hat{\mu}_{i, j}, \mathbf{I}\right)$:
\begin{equation}
\begin{split}
\mathcal{L}_{k l} &=K L\left[N\left(\mathbf{z}_{i,j} \mid \boldsymbol{\mu}_{i,j}, \boldsymbol{\sigma}_{i,j}^{2}\right) \| N(\mathbf{\hat{z}}_{i,j} \mid \boldsymbol{\hat{\mu}}_{i,j}, \mathbf{I})\right],
\label{depth kl}
\end{split}
\end{equation}
where $\boldsymbol{\hat{\mu}}_{i,j}$ is the ground truth depth value of the patch $\mathbf{x}_{i,j}$.
\par
\textbf{Training for FAS.}
When training the FAS model, each input image with a size of $3 \times 256 \times 256$ is first fed into a CNN-based depth feature extractor to obtain the depth feature map with the size of $64 \times 32 \times 32$. Specifically, in experiments, we use modified ResNet~\cite{He2016DeepRL}, MobileNetV2~\cite{Sandler2018MobileNetV2IR}, DepthNet~\cite{Liu2018learning} and CDCN~\cite{Yu2020SearchingCD} as the depth feature extractor to evaluate the universality of our method. Then, DUM is employed to transform the depth feature map to the predicted depth value $\mathbf{D}$. The architecture details are listed in Tab.~\ref{resandmb} and Tab.~\ref{dnandcdcn}. Finally, the mean square error loss $\mathcal{L}_{MSE}$ is utilized as the pixel-wise depth supervision constraint.
\par
The total objective function is:
\begin{equation}
\begin{split}
\mathcal{L}_{overall}=\mathcal{L}_{\text {MSE}}+\lambda_{k l} \mathcal{L}_{k l}+\lambda_{g} (\mathcal{L}^{'}_{MSE}+\lambda_{k l} \mathcal{L}^{'}_{k l}),
\label{ov loss}
\end{split}
\end{equation}
where $\mathcal{L}_{MSE}$ and $\mathcal{L}_{k l}$ represent the losses of the real data, and $\mathcal{L}^{'}_{MSE}$ and $\mathcal{L}^{'}_{k l}$ are the losses of the generated data. $\lambda_{k l}$ and $\lambda_{g}$ are the trade-off parameters. The former is adopted to control the regularization term, and the latter controls the proportion of effects caused by the generated data during backpropagation. Besides, in order to properly utilize the generated data to promote the training of FAS model, we construct each training batch by a combination of the original and the generated data with a ratio of $r$.
\section{Experiments}
\subsection{Datasets and Metrics}
\textbf{Datasets.}
Five FAS benchmarks are adopted for experiments, including OULU-NPU~\cite{Boulkenafet2017OULUNPUAM}, SiW~\cite{Liu2018learning}, SiW-M~\cite{Liu2019DeepTL}, CASIA-MFSD~\cite{Zhang2012CASIA} and Replay-Attack~\cite{Chingovska2012Replay}. OULU-NPU consists of 4,950 genuine and attack videos from 55 subjects. The attack videos contain two print attacks and two video replay attacks. There are four protocols to validate the generalization ability of models across different attacks, scenarios and video recorders. SiW contains 165 subjects with two print attacks and four video replay attacks. Three protocols evaluate the generalization ability with different poses and expressions, cross mediums and unknown attack types. SiW-M includes 13 attacks types (\textit{e.g.} print, replay, 3D mask, makeup and partial attacks) and more identities, which is usually employed for diverse spoof attacks evaluation. CASIA-MFSD and Replay-Attack are small-scale datasets contain low-resolution videos with photo and video attacks. Specifically, high-resolution dataset OULU-NPU and SiW are utilized to evaluate the intra-testing performance. For inter-testing, we conduct cross-type testing on the SiW-M dataset and cross-dataset testing between CASIA-MFSD and Replay-Attack.
\par
\textbf{Metrics.}
Our method is evaluated by the following metrics: Attack Presentation Classification Error Rate $(A P C E R)$, Bona Fide Presentation Classification Error Rate $(B P C E R)$ and Average Classification Error Rate $(A C E R)$:
\begin{equation}
A P C E R=\frac{F P}{F P+T N},
\end{equation}
\begin{equation}
B P C E R=\frac{F N}{F N+T P},
\end{equation}
\begin{equation}
A C E R=\frac{A P C E R+B P C E R}{2},
\end{equation}
where TP, TN, FP and FN denote True Positive, True Negative, False Positive and False Negative, respectively. In addition, following~\cite{Liu2018learning}, Equal Error Rate (EER) is additionally employed for cross-type evaluation, and Half Total Error Rate $(H T E R)$ is adopted in cross-dataset testing.
\subsection{Implementation Details}
We implement our method in PyTorch. In data generation phase, we use the same encoder and decoder networks as~\cite{Kingma2014AutoEncodingVB}. The learning rate is set to 2e-4 with Adam optimizer, and $\lambda_{1}$, $\lambda_{2}$, $\lambda_{3}$ and $\lambda_{4}$ in Eq.~\ref{gen loss} are empirically fixed with 50, 5, 1, 10, respectively. During training, we comply with the principle of not applying additional data. In face anti-spoofing phase, following~\cite{Yu2020SearchingCD, Wang2020DeepSG}, we employ PRNet~\cite{Feng2018Joint3F} to obtain the depth map of live face with a size of $32 \times 32$ and normalize the values within [0,1]. For the spoofing face, we set the depth map to zeros with the same size of 32x32. During training, we adopt the Adam optimizer with the initial learning rate of 1e-3. The trade-off parameters $\lambda_{k l}$ and $\lambda_{g}$ in Eq.~\ref{ov loss} are empirically set to 1e-3 and 1e-1, respectively, and the ratio $r$ is set to 0.75. We generate 20,000 images pairs by DSDG as the external data. During inference, we follow the routine of CDCN~\cite{Yu2020SearchingCD} to obtain the final score by averaging the predicted values in the depth map.
\begin{table}[t]
\scriptsize
\caption{Influence of different identity numbers.}
\vspace{-0.5em}
\begin{center}
\resizebox{.39\textwidth}{!}{
\begin{tabular}{c|cccc}
\hline
ID Number&5&10&15&20\\
\hline
CDCN~\cite{Yu2020SearchingCD}&8.3&4.3&1.9&1.0\\
\hline
Ours&5.1&2.2&0.9&\textbf{0.3}\\
\hline
\end{tabular}}
\end{center}
\label{identity num}
\vspace{0em}
\end{table}
\begin{table}[t]
\scriptsize
\caption{Ablation study of ratio $r$ on OULU-NPU P1.}
\vspace{-0.5em}
\begin{center}
\resizebox{.43\textwidth}{!}{
\begin{tabular}{c|ccccc}
\hline
ratio $r$&0&0.25&0.5&0.75&1\\
\hline
ACER(\%)&4.2&0.9&0.6&\textbf{0.3}&0.7\\
\hline
\end{tabular}}
\end{center}
\label{r ab}
\vspace{-0.5em}
\end{table}
\begin{table*}
\caption{Quantitative ablation study of the components in our method on OULU-NPU.}
\vspace{-0.5em}
\begin{center}
\resizebox{\textwidth}{!}{
\begin{tabular}{|c|c|c|c|c|c|c|c|c|}
\hline
Prot.&Method&ACER($\%$)&Method&ACER($\%$)&Method&ACER($\%$)&Method&ACER($\%$)\\
\hline
\hline
\multirow{4}{*}{1}
&ResNet~\cite{He2016DeepRL}&6.8&MobileNetV2~\cite{Sandler2018MobileNetV2IR}&5.0&DepthNet~\cite{Liu2018learning}&1.6&CDCN~\cite{Yu2020SearchingCD}&1.0\\
\cline{2-9}&ResNet+G&7.6($\uparrow$0.8)&MobileNetV2+G&4.6($\downarrow$0.4)&DepthNet+G&1.5($\downarrow$0.1)&CDCN+G&0.6($\downarrow$0.4)\\
\cline{2-9}&ResNet+U&5.0($\downarrow$1.8)&MobileNetV2+U&3.9($\downarrow$1.1)&DepthNet+U&1.5($\downarrow$0.1)&CDCN+U&0.7($\downarrow$0.3)\\
\cline{2-9}&ResNet+G+U&\textbf{4.8}($\downarrow$2.0)&MobileNetV2+G+U&\textbf{2.6}($\downarrow$2.4)&DepthNet+G+U&\textbf{1.3}($\downarrow$0.2)&CDCN+G+U&\textbf{0.3}($\downarrow$0.7)\\
\hline
\hline
\multirow{4}{*}{2}
&ResNet&3.1&MobileNetV2&1.9&DepthNet&2.7&CDCN&1.5\\
\cline{2-9}&ResNet+G&3.0($\downarrow$0.1)&MobileNetV2+G&1.7($\downarrow$0.2)&DepthNet+G&2.5($\downarrow$0.2)&CDCN+G&1.7($\uparrow$0.2)\\
\cline{2-9}&ResNet+U&3.0($\downarrow$0.1)&MobileNetV2+U&1.9&DepthNet+U&1.9($\downarrow$0.8)&CDCN+U&1.5\\
\cline{2-9}&ResNet+G+U&\textbf{2.8}($\downarrow$0.3)&MobileNetV2+G+U&\textbf{1.5}($\downarrow$0.4)&DepthNet+G+U&\textbf{1.8}($\downarrow$0.9)&CDCN+G+U&\textbf{1.2}($\downarrow$0.3)\\
\hline
\hline
\multirow{4}{*}{3}
&ResNet&$4.5\pm6.2$&MobileNetV2&$1.6\pm1.8$&DepthNet&$2.9\pm1.5$&CDCN&$2.3\pm1.4$\\
\cline{2-9}&ResNet+G&$3.6(\downarrow0.9)\pm5.7$&MobileNetV2+G&$2.2(\uparrow0.6)\pm3.6$&DepthNet+G&$2.3(\downarrow0.6)\pm3.4$&CDCN+G&$2.5(\uparrow0.2)\pm2.7$\\
\cline{2-9}&ResNet+U&$3.1(\downarrow1.4)\pm4.8$&MobileNetV2+U&$1.5(\downarrow0.1)\pm1.8$&DepthNet+U&$2.3(\downarrow0.6)\pm3.4$&CDCN+U&$2.2(\downarrow0.1)\pm2.8$\\
\cline{2-9}&ResNet+G+U&$\textbf{2.9}(\downarrow1.6)\pm\textbf{3.7}$&MobileNetV2+G+U&$\textbf{1.3}(\downarrow0.3)\pm\textbf{1.5}$&DepthNet+G+U&$\textbf{2.2}(\downarrow0.7)\pm\textbf{3.5}$&CDCN+G+U&$\textbf{1.4}(\downarrow0.9)\pm\textbf{1.5}$\\
\hline
\hline
\multirow{4}{*}{4}
&ResNet&$9.0\pm6.5$&MobileNetV2&$6.1\pm2.6$&DepthNet&$9.5\pm6.0$&CDCN&$6.9\pm2.9$\\
\cline{2-9}&ResNet+G&$6.4(\downarrow2.6)\pm8.0$&MobileNetV2+G&$4.8(\downarrow1.3)\pm4.4$&DepthNet+G&$4.4(\downarrow5.1)\pm3.6$&CDCN+G&$3.2(\downarrow3.7)\pm2.6$\\
\cline{2-9}&ResNet+U&$7.3(\downarrow2.7)\pm6.1$&MobileNetV2+U&$5.2(\downarrow0.9)\pm2.4$&DepthNet+U&$4.4(\downarrow5.1)\pm2.2$&CDCN+U&$3.2(\downarrow3.7)\pm1.3$\\
\cline{2-9}&ResNet+G+U&$\textbf{5.8}(\downarrow3.2)\pm\textbf{4.7}$&MobileNetV2+G+U&$\textbf{4.4}(\downarrow1.7)\pm\textbf{2.4}$&DepthNet+G+U&$\textbf{3.3}(\downarrow6.2)\pm\textbf{2.0}$&CDCN+G+U&$\textbf{2.3}(\downarrow4.6)\pm\textbf{2.3}$\\
\hline
\end{tabular}}
\end{center}
\vspace{0em}
\label{DUM DSDG ab}
\end{table*}
\begin{table}[t]
\scriptsize
\caption{Ablation study of the number of generated pairs.}
\vspace{-0.5em}
\begin{center}
\resizebox{.43\textwidth}{!}{
\begin{tabular}{c|ccccc}
\hline
Number(k)&10&15&20&25&30\\
\hline
ACER(\%)&0.7&0.6&\textbf{0.3}&\textbf{0.3}&\textbf{0.3}\\
\hline
\end{tabular}}
\end{center}
\label{num ab}
\vspace{0em}
\end{table}
\begin{table}[t]
\scriptsize
\caption{Ablation study of the losses related to identity learning on OULU-NPU P1.}
\vspace{-0.5em}
\begin{center}
\resizebox{.47\textwidth}{!}{
\begin{tabular}{c|cccc}
\hline
Methods&w$/$o $\mathcal{L}_{\text {mmd}}$&w$/$o $\mathcal{L}_{\text {pair}}$&w$/$o $\mathcal{L}_{\text {ort}}$&Ours\\
\hline
ACER(\%)&1.0&0.7&0.7&\textbf{0.3}\\
\hline
\end{tabular}}
\end{center}
\label{identity loss ab}
\vspace{-0.5em}
\end{table}
\begin{table}[t]
\scriptsize
\caption{Analysis of $\lambda_{g}$ and $\lambda_{kl}$ on OULU-NPU P1.}
\vspace{-0.5em}
\begin{center}
\resizebox{.43\textwidth}{!}{
\begin{tabular}{c|ccccc}
\hline
$\lambda_{g}$&0.2&0.1&0.05&0.02&0.01\\
\hline
ACER(\%)&0.9&\textbf{0.3}&0.8&0.9&1.4\\
\hline
\hline
$\lambda_{kl}$&1&1e-1&1e-2&1e-3&1e-4\\
\hline
ACER(\%)&0.7&0.7&0.6&\textbf{0.3}&0.8\\
\hline
\end{tabular}}
\end{center}
\label{lambda ab}
\vspace{0em}
\end{table}
\begin{table}[t]
\scriptsize
\caption{Influence of different number of unknown spoof types on OULU-NPU P1.}
\vspace{-0.5em}
\begin{center}
\resizebox{.43\textwidth}{!}{
\begin{tabular}{c|ccccc}
\hline
Number&0&1&2&3&4\\
\hline
ACER(\%)&\textbf{0.3}&0.5&0.6&0.6&0.7\\
\hline
\end{tabular}}
\end{center}
\label{unknowlabel}
\vspace{-1em}
\end{table}
\subsection{Ablation Study}
\label{ablation study}
\textbf{Influence of the identity number.}
Tab.~\ref{identity num} presents the influence of different identity numbers to the model performance. We choose 5, 10, 15 and 20 face identities from OULU-NPU P1 as the training data. It is obvious that as the identity number increasing, the performance of the model gets better. Besides, as shown in Tab.~\ref{identity num}, our method obtains better performance than the raw CDCN on each setting, especially when the number of identity is insufficient. We argue that more face identities cover more facial features and facial structures, which can significantly improve the model generalization ability.
\par
\textbf{Influence of the ratio between the original data and the generated data in a batch.}
The hyper-parameter $r$ controls the ratio of the original data over the generated data in each training batch. Specifically, $r$=1 or $r$=0 represents only using the original data or the generated data, respectively. We generate 20,000 images pairs and vary the ratio $r$ on OULU-NPU P1. As shown in Tab.~\ref{r ab}, with a proper value of ratio $r$ ($r$=0.5 or $r$=0.75), the model achieves better performance than only using the original data ($r$=1), indicating the generated data can promote the training process, when $r$ is equal to 0.75, the model obtains the best result. Then, we fix the $r$ to 0.75 and gradually increase the generated image pairs from 10,000 to 30,000 with 5,000 intervals. As shown in Tab.~\ref{num ab}, the ACER is 0.7, 0.6. 0.3, 0.3 and 0.3, respectively. Thus, if not specially indicated, we fix the $r$ to 0.75 and generate 20,000 images pairs for all experiments.
\par
\begin{table}
\caption{The results of intra-testing on OULU-NPU.}
\vspace{-0.5em}
\begin{center}
\resizebox{\columnwidth}{!}{
\begin{tabular}{|c|c|c|c|c|}
\hline
Prot.&Method&APCER($\%$)&BPCER($\%$)&ACER($\%$)\\
\hline
\hline
\multirow{14}{*}{1}
&DRL-FAS~\cite{Cai2021DRLFASAN}
&5.4&4.0&4.7\\
\cline{2-5}&CIFL~\cite{Chen2021CameraIF} 
&3.8&2.9&3.4\\
\cline{2-5}&Auxiliary~\cite{Liu2018learning} 
&1.6&1.6&1.6\\
\cline{2-5}&MA-Net~\cite{Liu2021FaceAV}
&1.4&1.8&1.6\\
\cline{2-5}&DENet~\cite{Zheng2021AttentionBasedSM}
&1.7&0.8&1.3\\
\cline{2-5}&Disentangled~\cite{Zhang2020FaceAV} 
&1.7&0.8&1.3\\
\cline{2-5}&SpoofTrace~\cite{Liu2020OnDS}
&0.8&1.3&1.1\\
\cline{2-5}&FAS-SGTD~\cite{Wang2020DeepSG} 
&2.0&0&1.0\\
\cline{2-5}&CDCN~\cite{Yu2020SearchingCD} 
&0.4&1.7&1.0\\
\cline{2-5}&BCN~\cite{Yu2020FaceAW}
&0&1.6&0.8\\
\cline{2-5}&CDCN-PS~\cite{Yu2021RevisitingPS}
&0.4&1.2&0.8\\
\cline{2-5}&SCNN-PL-TC~\cite{Quan2021ProgressiveTL}
&$0.6\pm0.4$&$0.0\pm0.0$&$0.4\pm0.2$\\
\cline{2-5}&DC-CDN~\cite{Yu2021DualCrossCD}
&0.5&0.3&0.4\\
\cline{2-5}&\textbf{Ours}
&0.6&0.0&\textbf{0.3}\\
\hline
\hline
\multirow{14}{*}{2}
&Auxiliary~\cite{Liu2018learning} 
&2.7&2.7&2.7\\
\cline{2-5}&MA-Net~\cite{Liu2021FaceAV}
&4.4&0.6&2.5\\
\cline{2-5}&CIFL~\cite{Chen2021CameraIF} 
&3.6&1.2&2.4\\
\cline{2-5}&Disentangled~\cite{Zhang2020FaceAV} 
&1.1&3.6&2.4\\
\cline{2-5}&DRL-FAS~\cite{Cai2021DRLFASAN}
&3.7&0.1&1.9\\
\cline{2-5}&FAS-SGTD~\cite{Wang2020DeepSG} 
&2.5&1.3&1.9\\
\cline{2-5}&SpoofTrace~\cite{Liu2020OnDS}
&2.3&1.6&1.9\\
\cline{2-5}&BCN~\cite{Yu2020FaceAW}
&2.6&0.8&1.7\\
\cline{2-5}&DENet~\cite{Zheng2021AttentionBasedSM}
&2.6&0.8&1.7\\
\cline{2-5}&CDCN~\cite{Yu2020SearchingCD} 
&1.5&1.4&1.5\\
\cline{2-5}&CDCN-PS~\cite{Yu2021RevisitingPS}
&1.4&1.4&1.4\\
\cline{2-5}&DC-CDN~\cite{Yu2021DualCrossCD}
&0.7&1.9&1.3\\
\cline{2-5}&SCNN-PL-TC~\cite{Quan2021ProgressiveTL}
&$1.7\pm0.9$&$0.6\pm0.3$&$1.2\pm0.5$\\
\cline{2-5}&\textbf{Ours}
&1.5&0.8&\textbf{1.2}\\
\hline
\hline
\multirow{14}{*}{3}
&DRL-FAS~\cite{Cai2021DRLFASAN}
&$4.6\pm1.3$&$1.3\pm1.8$&$3.0\pm1.5$\\
\cline{2-5}&Auxiliary~\cite{Liu2018learning} 
&$2.7\pm1.3$&$3.1\pm1.7$&$2.9\pm1.5$\\
\cline{2-5}&SpoofTrace~\cite{Liu2020OnDS}
&$1.6\pm1.6$&$4.0\pm5.4$&$2.8\pm3.3$\\
\cline{2-5}&DENet~\cite{Zheng2021AttentionBasedSM}
&$2.0\pm2.6$&$3.9\pm2.2$&$2.8\pm2.4$\\
\cline{2-5}&FAS-SGTD~\cite{Wang2020DeepSG} 
&$3.2\pm2.0$&$2.2\pm1.4$&$2.7\pm0.6$\\
\cline{2-5}&BCN~\cite{Yu2020FaceAW}
&$3.2\pm2.0$&$2.2\pm1.4$&$2.7\pm0.6$\\
\cline{2-5}&CIFL~\cite{Chen2021CameraIF}
&$3.8\pm1.1$&$1.1\pm1.1$&$2.5\pm0.8$\\
\cline{2-5}&CDCN~\cite{Yu2020SearchingCD} 
&$2.4\pm1.3$&$2.2\pm2.0$&$2.3\pm1.4$\\
\cline{2-5}&Disentangled~\cite{Zhang2020FaceAV} 
&$2.8\pm2.2$&$1.7\pm2.6$&$2.2\pm2.2$\\
\cline{2-5}&CDCN-PS~\cite{Yu2021RevisitingPS}
&$1.9\pm1.7$&$2.0\pm1.8$&$2.0\pm1.7$\\
\cline{2-5}&DC-CDN~\cite{Yu2021DualCrossCD}
&$2.2\pm2.8$&$1.6\pm2.1$&$1.9\pm1.1$\\
\cline{2-5}&SCNN-PL-TC~\cite{Quan2021ProgressiveTL}
&$1.5\pm0.9$&$2.2\pm1.0$&$1.7\pm0.8$\\
\cline{2-5}&MA-Net~\cite{Liu2021FaceAV}
&$1.5\pm1.2$&$1.6\pm1.1$&$1.6\pm1.1$\\
\cline{2-5}&\textbf{Ours}
&$1.2\pm0.8$&$1.7\pm3.3$&$\textbf{1.4}\pm\textbf{1.5}$\\
\hline
\hline
\multirow{14}{*}{4}
&Auxiliary~\cite{Liu2018learning} 
&$9.3\pm5.6$&$10.4\pm6.0$&$9.5\pm6.0$\\
\cline{2-5}&DRL-FAS~\cite{Cai2021DRLFASAN}
&$8.1\pm2.7$&$6.9\pm5.8$&$7.2\pm3.9$\\
\cline{2-5}&CDCN~\cite{Yu2020SearchingCD} 
&$4.6\pm4.6$&$9.2\pm8.0$&$6.9\pm2.9$\\
\cline{2-5}&CIFL~\cite{Chen2021CameraIF}
&$5.9\pm3.3$&$6.3\pm4.7$&$6.1\pm4.1$\\
\cline{2-5}&MA-Net~\cite{Liu2021FaceAV}
&$5.4\pm3.2$&$5.5\pm2.8$&$5.5\pm3.6$\\
\cline{2-5}&BCN~\cite{Yu2020FaceAW}
&$2.9\pm4.0$&$7.5\pm6.9$&$5.2\pm3.7$\\
\cline{2-5}&FAS-SGTD~\cite{Wang2020DeepSG} 
&$6.7\pm7.5$&$3.3\pm4.1$&$5.0\pm2.2$\\
\cline{2-5}&SCNN-PL-TC~\cite{Quan2021ProgressiveTL}
&$5.2\pm2.0$&$4.6\pm4.1$&$4.8\pm2.0$\\
\cline{2-5}&CDCN-PS~\cite{Yu2021RevisitingPS}
&$2.9\pm4.0$&$5.8\pm4.9$&$4.8\pm1.8$\\
\cline{2-5}&DENet~\cite{Zheng2021AttentionBasedSM}
&$4.2\pm5.2$&$4.6\pm3.8$&$4.4\pm4.5$\\
\cline{2-5}&Disentangled~\cite{Zhang2020FaceAV}
&$5.4\pm2.9$&$3.3\pm6.0$&$4.4\pm3.0$\\
\cline{2-5}&DC-CDN~\cite{Yu2021DualCrossCD}
&$5.4\pm3.3$&$2.5\pm4.2$&$4.0\pm3.1$\\
\cline{2-5}&SpoofTrace~\cite{Liu2020OnDS}
&$2.3\pm3.6$&$5.2\pm5.4$&$3.8\pm4.2$\\
\cline{2-5}&\textbf{Ours}
&$2.1\pm1.0$&$2.5\pm4.2$&$\textbf{2.3}\pm\textbf{2.3}$\\
\hline
\end{tabular}}
\end{center}
\vspace{-1em}
\label{OULU-NPU}
\end{table}
\begin{table}
\caption{The results of intra-testing on SiW.}
\vspace{-0.5em}
\begin{center}
\resizebox{\columnwidth}{!}{
\begin{tabular}{|c|c|c|c|c|}
\hline
Prot.&Method&APCER($\%$)&BPCER($\%$)&ACER($\%$)\\
\hline
\hline
\multirow{10}{*}{1}
&Auxiliary~\cite{Liu2018learning} 
&3.58&3.58&3.58\\
\cline{2-5}&MA-Net~\cite{Liu2021FaceAV} 
&0.41&1.01&0.71\\
\cline{2-5}&FAS-SGTD~\cite{Wang2020DeepSG} 
&0.64&0.17&0.40\\
\cline{2-5}&BCN~\cite{Yu2020FaceAW}
&0.55&0.17&0.36\\
\cline{2-5}&Disentangled~\cite{Zhang2020FaceAV} 
&0.07&0.50&0.28\\
\cline{2-5}&CDCN~\cite{Yu2020SearchingCD} 
&0.07&0.17&0.12\\
\cline{2-5}&SpoofTrace~\cite{Liu2020OnDS}
&0.00&0.00&\textbf{0.00}\\
\cline{2-5}&DRL-FAS~\cite{Cai2021DRLFASAN}
&-&-&\textbf{0.00}\\
\cline{2-5}&DENet~\cite{Zheng2021AttentionBasedSM}
&0.00&0.00&\textbf{0.00}\\
\cline{2-5}&\textbf{Ours}
&0.00&0.00&\textbf{0.00}\\
\hline
\hline
\multirow{10}{*}{2}
&Auxiliary~\cite{Liu2018learning} 
&$0.57\pm0.69$&$0.57\pm0.69$&$0.57\pm0.69$\\
\cline{2-5}&MA-Net~\cite{Liu2021FaceAV} 
&$0.19\pm0.25$&$0.64\pm0.97$&$0.42\pm0.34$\\
\cline{2-5}&BCN~\cite{Yu2020FaceAW}
&$0.08\pm0.17$&$0.15\pm0.00$&$0.11\pm0.08$\\
\cline{2-5}&Disentangled~\cite{Zhang2020FaceAV} 
&$0.08\pm0.17$&$0.13\pm0.09$&$0.10\pm0.04$\\
\cline{2-5}&CDCN~\cite{Yu2020SearchingCD} 
&$0.00\pm0.00$&$0.09\pm0.10$&$0.04\pm0.05$\\
\cline{2-5}&FAS-SGTD~\cite{Wang2020DeepSG} 
&$0.00\pm0.00$&$0.04\pm0.08$&$0.02\pm0.04$\\
\cline{2-5}&SpoofTrace~\cite{Liu2020OnDS}
&$0.00\pm0.00$&$0.00\pm0.00$&$\textbf{0.00}\pm\textbf{0.00}$\\
\cline{2-5}&DRL-FAS~\cite{Cai2021DRLFASAN}
&-&-&$\textbf{0.00}\pm\textbf{0.00}$\\
\cline{2-5}&DENet~\cite{Zheng2021AttentionBasedSM}
&$0.00\pm0.00$&$0.00\pm0.00$&$\textbf{0.00}\pm\textbf{0.00}$\\
\cline{2-5}&\textbf{Ours}
&$0.00\pm0.00$&$0.00\pm0.00$&$\textbf{0.00}\pm\textbf{0.00}$\\
\hline
\hline
\multirow{10}{*}{3}
&Auxiliary~\cite{Liu2018learning} 
&$8.31\pm3.81$&$8.31\pm3.8$&$8.31\pm3.81$\\
\cline{2-5}&SpoofTrace~\cite{Liu2020OnDS}
&$8.30\pm3.30$&$7.50\pm3.30$&$7.90\pm3.30$\\
\cline{2-5}&Disentangled~\cite{Zhang2020FaceAV} 
&$9.35\pm6.14$&$1.84\pm2.60$&$5.59\pm4.37$\\
\cline{2-5}&DRL-FAS~\cite{Cai2021DRLFASAN}
&-&-&$4.51\pm0.00$\\
\cline{2-5}&DENet~\cite{Zheng2021AttentionBasedSM}
&$3.74\pm0.54$&$4.97\pm0.64$&$4.35\pm0.59$\\
\cline{2-5}&MA-Net~\cite{Liu2021FaceAV} 
&$4.21\pm3.22$&$4.47\pm2.63$&$4.34\pm3.29$\\
\cline{2-5}&FAS-SGTD~\cite{Wang2020DeepSG} 
&$2.63\pm3.72$&$2.92\pm3.42$&$2.78\pm3.57$\\
\cline{2-5}&BCN~\cite{Yu2020FaceAW}
&$2.55\pm0.89$&$2.34\pm0.47$&$2.45\pm0.68$\\
\cline{2-5}&CDCN~\cite{Yu2020SearchingCD} 
&$1.67\pm0.11$&$1.76\pm0.12$&$\textbf{1.71}\pm\textbf{0.11}$\\
\cline{2-5}&\textbf{Ours}
&$3.75\pm1.46$&$3.85\pm1.42$&$3.80\pm1.44$\\
\hline
\end{tabular}}
\end{center}
\vspace{0em}
\label{SIW}
\end{table}
\begin{table}
\caption{The results of cross-dataset testing between CASIA-MFSD and Replay-Attack. The evaluation metric is HTER (\%).}
\vspace{-0.5em}
\begin{center}
\resizebox{\columnwidth}{!}{
\begin{tabular}{|c|c|c|c|c|}
\hline
\multirow{3}{*}{Method}&
Train&Test&Train&Test\\
\cline{2-5}
&CASIA-&Replay-&Replay-&CASIA-\\
&MFSD&Attack&Attack&MFSD\\
\hline
LBP~\cite{zine2015face} 
&\multicolumn{2}{c|}{47.0}&\multicolumn{2}{c|}{39.6}\\
\cline{1-5}STASN~\cite{Yang2019FaceAM} 
&\multicolumn{2}{c|}{31.5}&\multicolumn{2}{c|}{30.9}\\
\cline{1-5}FaceDe-S~\cite{Jourabloo2018FaceDA}
&\multicolumn{2}{c|}{28.5}&\multicolumn{2}{c|}{41.1}\\
\cline{1-5}DRL-FAS~\cite{Cai2021DRLFASAN}
&\multicolumn{2}{c|}{28.4}&\multicolumn{2}{c|}{33.2}\\
\cline{1-5}Auxiliary~\cite{Liu2018learning} 
&\multicolumn{2}{c|}{27.6}&\multicolumn{2}{c|}{28.4}\\
\cline{1-5}DENet~\cite{Zheng2021AttentionBasedSM}
&\multicolumn{2}{c|}{27.4}&\multicolumn{2}{c|}{28.1}\\
\cline{1-5}Disentangled~\cite{Zhang2020FaceAV} 
&\multicolumn{2}{c|}{22.4}&\multicolumn{2}{c|}{30.3}\\
\cline{1-5}FAS-SGTD~\cite{Wang2020DeepSG} 
&\multicolumn{2}{c|}{17.0}&\multicolumn{2}{c|}{\textbf{22.8}}\\
\cline{1-5}BCN~\cite{Yu2020FaceAW} 
&\multicolumn{2}{c|}{16.6}&\multicolumn{2}{c|}{36.4}\\
\cline{1-5}CDCN~\cite{Yu2020SearchingCD} 
&\multicolumn{2}{c|}{15.5}&\multicolumn{2}{c|}{32.6}\\
\cline{1-5}CDCN-PS~\cite{Yu2021RevisitingPS} 
&\multicolumn{2}{c|}{13.8}&\multicolumn{2}{c|}{31.3}\\
\cline{1-5}DC-CDN~\cite{Yu2021DualCrossCD} 
&\multicolumn{2}{c|}{\textbf{6.0}}&\multicolumn{2}{c|}{30.1}\\
\cline{1-5}\textbf{Ours}
&\multicolumn{2}{c|}{15.1}&\multicolumn{2}{c|}{26.7}\\
\hline
\end{tabular}}
\end{center}
\vspace{-1em}
\label{CASIA-Replay}
\end{table}
\textbf{Quantitative analysis of the identity learning.}
The $\mathcal{L}_{\text {mmd}}$, $\mathcal{L}_{\text {pair}}$ and $\mathcal{L}_{\text {ort}}$ are utilized to effectively disentangle the spoof images into the spoofing pattern representation and the identity representation. In order to investigate the effectiveness of each loss, during training DSDG, we discard $\mathcal{L}_{\text {mmd}}$, $\mathcal{L}_{\text {pair}}$ and $\mathcal{L}_{\text {ort}}$ in Eq.~\ref{gen loss}, and evaluate the performance on OULU-NPU P1, respectively. As shown in Tab.~\ref{identity loss ab}, it can be observed that the performance drops significantly if one of the losses is discarded, which further demonstrates the importance of each identity loss.
\par
\textbf{Sensitivity analysis of $\lambda_{kl}$ and $\lambda_{g}$.}
Tab.~\ref{lambda ab} shows the analysis of sensitivity study for the hyper-parameters $\lambda_{g}$ and $\lambda_{kl}$ in Eq.~\ref{ov loss}, where $\lambda_{g}$ controls the proportion of effects caused by the generated data in backpropagation and $\lambda_{kl}$ is the trade-off parameter of the Kullback-Leibler divergence constraint. Specifically, when setting $\lambda_{g}$ and $\lambda_{kl}$ to 0.1 and 1e-3, respectively, the model achieves the best ACER. In this situation, we find all loss values fall into a similar magnitude. Besides, most of the results outperform the CDCN whose ACER is 1.0, indicating that our method is not sensitive to these trade-off parameters in a large range.
\par
\textbf{Effectiveness of the DSDG and DUM.}
Tab.~\ref{DUM DSDG ab} presents the ablation study of different components in our method. ``G'' and ``U'' are short for DSDG and DUM, respectively. Modified ResNet, MobileNetV2, DepthNet and CDCN are employed as the baseline. We conduct the ablation study on all of four protocols on OULU-NPU~\cite{Boulkenafet2017OULUNPUAM}. From the comparison results, it is obvious that combining DSDG and DUM can facilitate the models to obtain better generalization ability, especially on the challenging Protocol 4. On the other hand, both DSDG and DUM are universal methods that can be integrated into different network backbones. Meanwhile, we also find that the performance degrades on a few protocols when only adopt DSDG. We attribute this situation to the presence of noisy samples in the generated data, which is nevertheless solved by DUM. In other word, depth uncertainty learning can indeed handle the adverse effects of partial distortion in the generated images, even brings significant improvement in the case of only using the original data.
\par
\begin{table*}[htp]
\caption{The evaluation results of the cross-type testing on SiW-M.}
\vspace{-0.5em}
\begin{center}
\centering
\resizebox{\textwidth}{!}{
\begin{tabular}{|c|c|c|c|c|c|c|c|c|c|c|c|c|c|c|c|}
\hline
\multirow{2}{*}{Method}&
\multirow{2}{*}{Metrics($\%$)}&
\multirow{2}{*}{Replay}&
\multirow{2}{*}{Print}&
\multicolumn{5}{c|}{Mask Attacks}&
\multicolumn{3}{c|}{Makeup Attacks}&
\multicolumn{3}{c|}{Partial Attacks}&
\multirow{2}{*}{Average}\\
\cline{5-15}&&&&Half&Silic.&Trans.&Paper.&Manne.&Ob.&Imp.&Cos.&Funny Eye&Paper Gls.&Part. Paper&
\\
\hline\hline
\multirow{2}{*}{Auxiliary~\cite{Liu2018learning}}
&ACER
&16.8&6.9&19.3&14.9&52.1&8.0&12.8&55.8&13.7&11.7&49.0&40.5&5.3&$23.6\pm18.5$\\
\cline{2-16}&EER
&14.0&4.3&11.6&12.4&24.6&7.8&10.0&72.3&10.1&\textbf{9.4}&21.4&18.6&4.0&$12.0\pm10.0$\\
\hline\hline
\multirow{2}{*}{SpoofTrace\cite{Liu2020OnDS}}
&ACER
&7.8&7.3&7.1&12.9&13.9&4.3&6.7&53.2&4.6&19.5&20.7&21.0&5.6&$14.2\pm13.2$\\
\cline{2-16}&EER
&7.6&\textbf{3.8}&8.4&13.8&14.5&5.3&4.4&35.4&\textbf{0.0}&19.3&21.0&20.8&1.6&$12.0\pm10.0$\\
\hline\hline
\multirow{2}{*}{CDCN\cite{Yu2020SearchingCD}}
&ACER
&8.7&7.7&11.1&9.1&20.7&4.5&5.9&44.2&2.0&15.1&25.4&19.6&3.3&$13.6\pm11.7$\\
\cline{2-16}&EER
&8.2&7.8&8.3&7.4&20.5&5.9&5.0&47.8&1.6&14.0&24.5&18.3&1.1&$13.1\pm12.6$\\
\hline\hline
\multirow{2}{*}{CDCN-PS\cite{Yu2021RevisitingPS}}
&ACER
&12.1&7.4&9.9&9.1&14.8&5.3&5.9&43.1&\textbf{0.4}&13.8&24.4&18.1&3.5&$12.9\pm11.1$\\
\cline{2-16}&EER
&10.3&7.8&8.3&7.4&10.2&5.9&5.0&43.4&\textbf{0.0}&12.0&23.9&15.9&\textbf{0.0}&$11.5\pm11.4$\\
\hline\hline
\multirow{2}{*}{SSR-FCN\cite{Deb2021LookLI}}
&ACER
&\textbf{7.4}&19.5&\textbf{3.2}&7.7&33.3&5.2&3.3&\textbf{22.5}&5.9&11.7&21.7&\textbf{14.1}&6.4&$12.4\pm9.2$\\
\cline{2-16}&EER
&\textbf{6.8}&11.2&\textbf{2.8}&\textbf{6.3}&28.5&\textbf{0.4}&3.3&\textbf{17.8}&3.9&11.7&21.6&\textbf{13.5}&3.6&$10.1\pm8.4$\\
\hline\hline
\multirow{2}{*}{DC-CDN\cite{Yu2021DualCrossCD}}
&ACER
&12.1&9.7&14.1&\textbf{7.2}&14.8&4.5&\textbf{1.6}&40.1&\textbf{0.4}&\textbf{11.4}&\textbf{20.1}&16.1&2.9&$11.9\pm10.3$\\
\cline{2-16}&EER
&10.3&8.7&11.1&7.4&12.5&5.9&\textbf{0.0}&39.1&\textbf{0.0}&12.0&\textbf{18.9}&13.5&1.2&$10.8\pm10.1$\\
\hline\hline
\multirow{2}{*}{BCN\cite{Yu2020FaceAW}}
&ACER
&12.8&\textbf{5.7}&10.7&10.3&14.9&\textbf{1.9}&2.4&32.3&0.8&12.9&22.9&16.5&\textbf{1.7}&$11.2\pm9.2$\\
\cline{2-16}&EER
&13.4&5.2&8.3&9.7&13.6&5.8&2.5&33.8&\textbf{0.0}&14.0&23.3&16.6&1.2&$11.3\pm9.5$\\
\hline\hline
\multirow{2}{*}{\textbf{Ours}}
&ACER
&7.8&7.3&9.1&8.4&\textbf{12.4}&4.5&4.4&32.7&\textbf{0.4}&12.0&22.5&\textbf{14.1}&2.3&$\textbf{10.6}\pm\textbf{8.8}$\\
\cline{2-16}&EER
&7.1&6.9&4.2&7.4&\textbf{10.2}&5.9&2.5&30.4&\textbf{0.0}&14.0&20.1&15.1&\textbf{0.0}&$\textbf{9.5}\pm\textbf{8.6}$\\
\hline
\end{tabular}}
\end{center}
\label{SIW-M}
\vspace{0em}
\end{table*}
\textbf{Influence of different number of unknown spoof types.}
The spoof type label is used for better disentangling the specific spoof pattern during training of DSDG. In some cases, the spoof types of images may be unavailable. We design an experiment to explore the influence of different number of unknown spoof types. Specifically, there are four spoof types in OULU-NPU P1, we assume that some of the spoof type labels are unknown, and set them as a class of ``unknown" to train DSDG. We gradually increase the number of unknown spoof types from 0 to 4, where 4 means that all the spoof type labels are unavailable. The results are shown in Tab.~\ref{unknowlabel}. We can observe that ACER gets worse with more unknown spoof types, but in the worst case (Number = 4), our method still obtains 0.7 in ACER, which outperforms the previous best 0.8 in ACER and beats CDCN by 0.3 in ACER. Thus, our method can still promote the performance without the fine grained spoof type labels.
\par
\subsection{Intra Testing}
We implement the intra-testing on the OULU-NPU and SiW datasets. In order to ensure the fairness of the comparison, we split the data used for DSDG training according to the protocols of each dataset (\textit{e.g.} OULU-NPU and SiW own 14 and 7 sub-protocols, respectively), and employ CDCN~\cite{Yu2020SearchingCD} as the depth feature extractor, which is orthogonal to our contribution.
\par
\begin{figure*}[htp]
\centering
\includegraphics[scale=0.68]{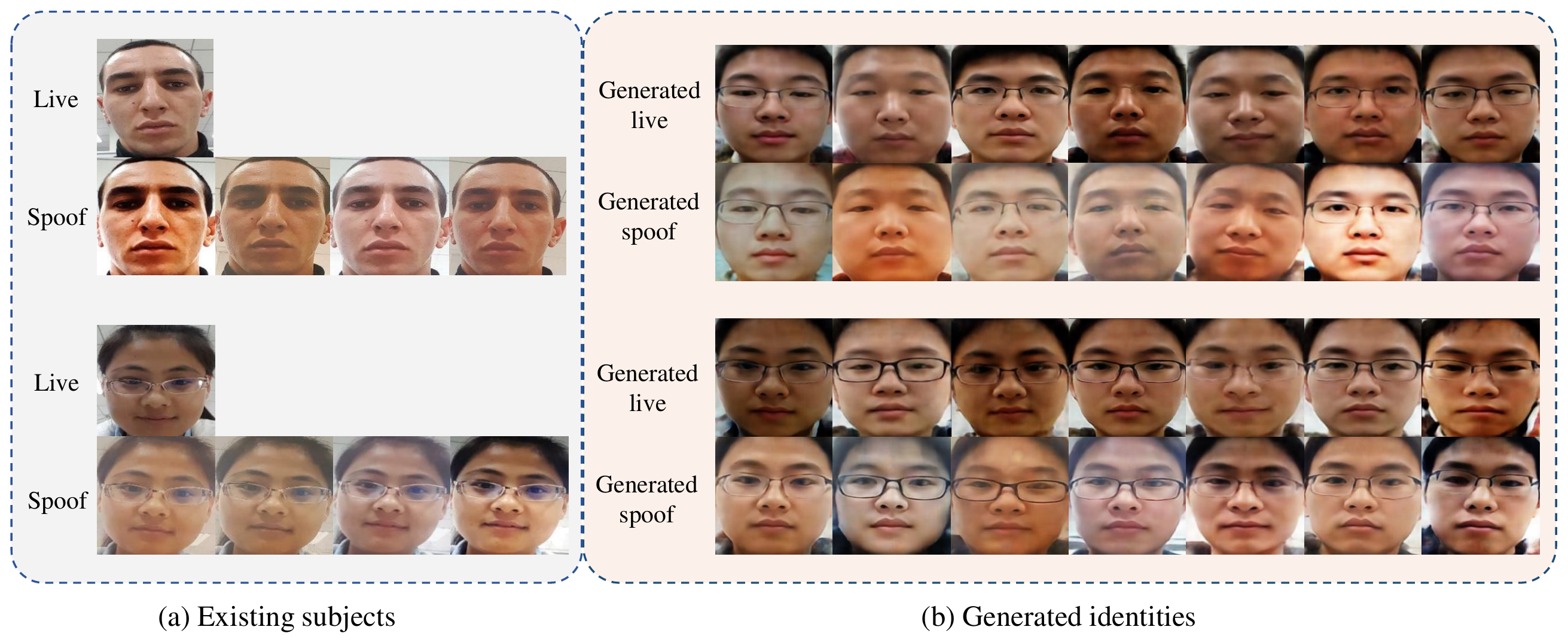}
\caption{Visualization results of DSDG on OULU-NPU Protocol 1. (a) The real paired live and spoofing images from two identities. The 2nd and 4th rows are spoofing images, containing four spoof types (\textit{e.g.} Print1, Print2, Replay1 and Replay2). Each column corresponds to a spoof type. (b) The generated paired live and spoofing images contain abundant identities. It can be seen that the generated spoofing images preserve the original spoof patterns of four spoof types.}
\label{render}
\vspace{-0em}
\end{figure*}
\textbf{Results on OULU-NPU.}
Tab.~\ref{OULU-NPU} shows the comparisons on OULU-NPU. Our proposed method obtains the best results on all the protocols. It is worth mentioning that the performance is significantly improved by our method on the most challenging Protocol 4 which focuses on the evaluation across unseen environmental conditions, attacks and input sensors. That means our method is able to obtain a more generic model by adopting a more diverse training set with depth uncertainty learning.
\par
\textbf{Results on SiW.}
Tab.~\ref{SIW} presents the results on SiW, where our method is compared with other state-of-the-art methods on three protocols. It can be seen that our method achieves the best performance on the first two protocols and a competitive performance on Protocol 3. Note that, we obtain the non-ideal result (4.34\%, 4.35\%, 4.34\% for APCER, BPCER, ACER, respectively) when reproduce the CDCN on Protocol 3. Thus, our method still has the capacity of improving the generalization ability while encounter unknown presentation attacks.
\subsection{Inter Testing}
\textbf{Cross-type Testing.}
SiW-M contains more diverse presentation attacks, and is more suitable for evaluating the generalization ability to unknown attacks. As shown in Tab.~\ref{SIW-M}, comparisons with seven state-of-the-art methods on leave-one-out protocols are conducted. Note that, unlike other datasets, the paired live and spoofing images are not accessible on SiW-M. Thus, we discard $\mathcal{L}_{\text {mmd}}$ and $\mathcal{L}_{\text {pair }}$ in Eq.~\ref{gen loss} when utilizing DSDG to generate data. In such case, our method still achieves the best average ACER(10.6\%) and EER(9.5\%), outperforming all the previous methods, as well as the best EER and ACER against most of the 13 attacks. Particularly, our method yields a significant improvement (3.0\% ACER and 3.6\% EER) compared with CDCN, benefiting from the advantages of DSDG and DUM.
\par
\begin{figure*}
\centering
\includegraphics[scale=0.64]{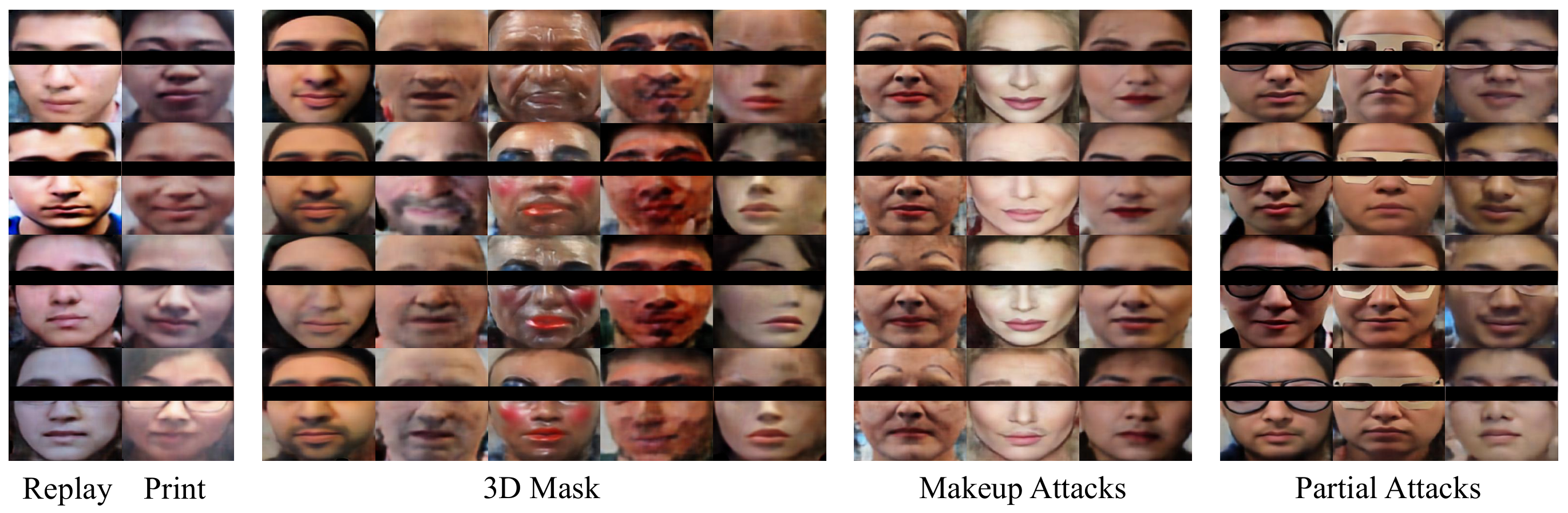}
\caption{
Visualization results of DSDG for diverse presentation attacks on SiW-M. It can be seen that DSDG generates high quality spoofing images with diverse identities and diverse attack types, which retain the original spoofing patterns.
}
\label{render-siw}
\vspace{-0.5em}
\end{figure*}
\begin{figure}[t]
\centering
\includegraphics[scale=0.62]{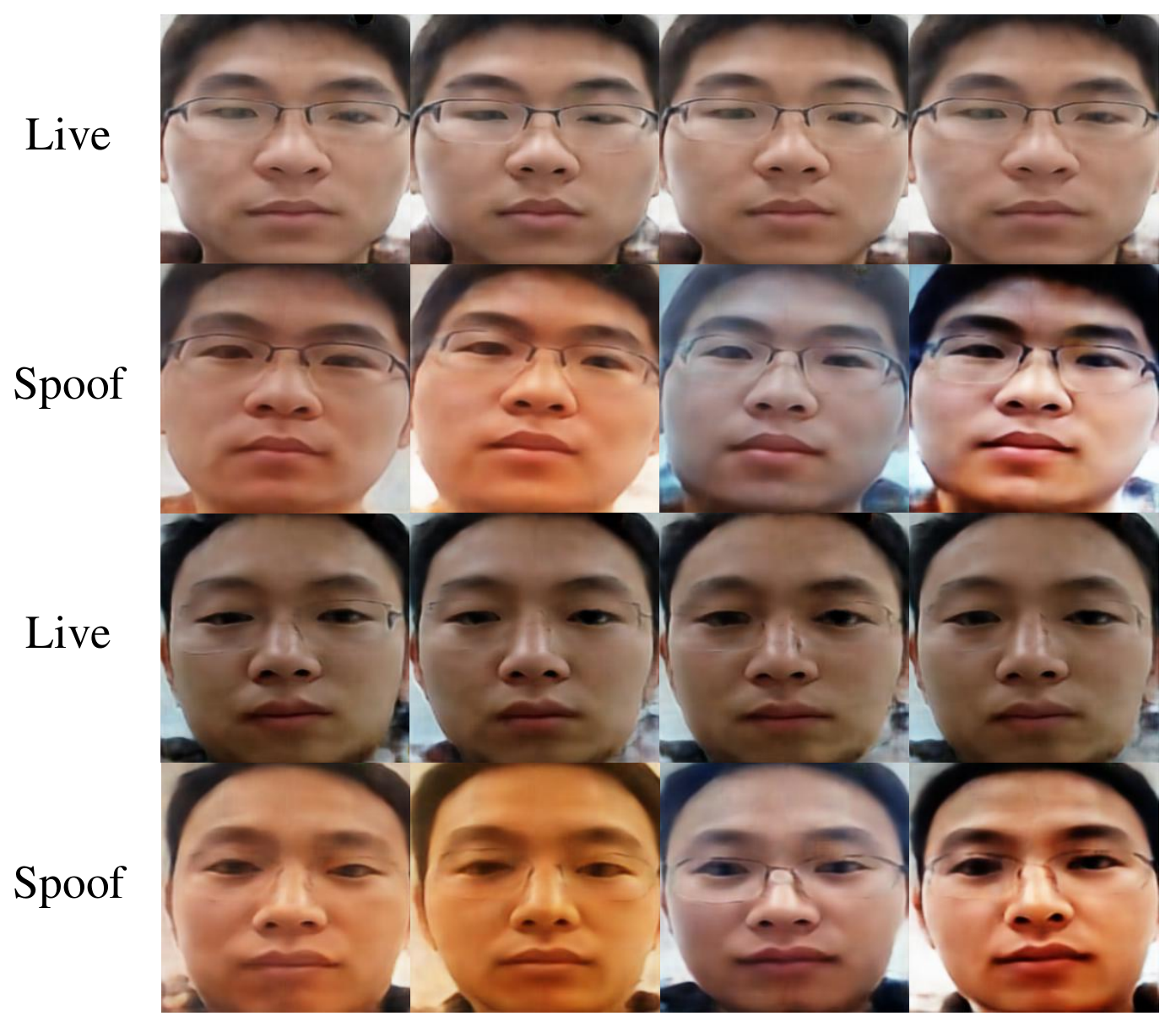}
\caption{
Generated paired live and spoofing faces by DSDG on the OULU-NPU Protocol 1. The odd rows are live images with the same  identity, and the even rows contain spoofing images with various spoof types.
}
\label{render-oulu}
\vspace{-0em}
\end{figure}
\begin{figure}
\centering
\includegraphics[scale=0.58]{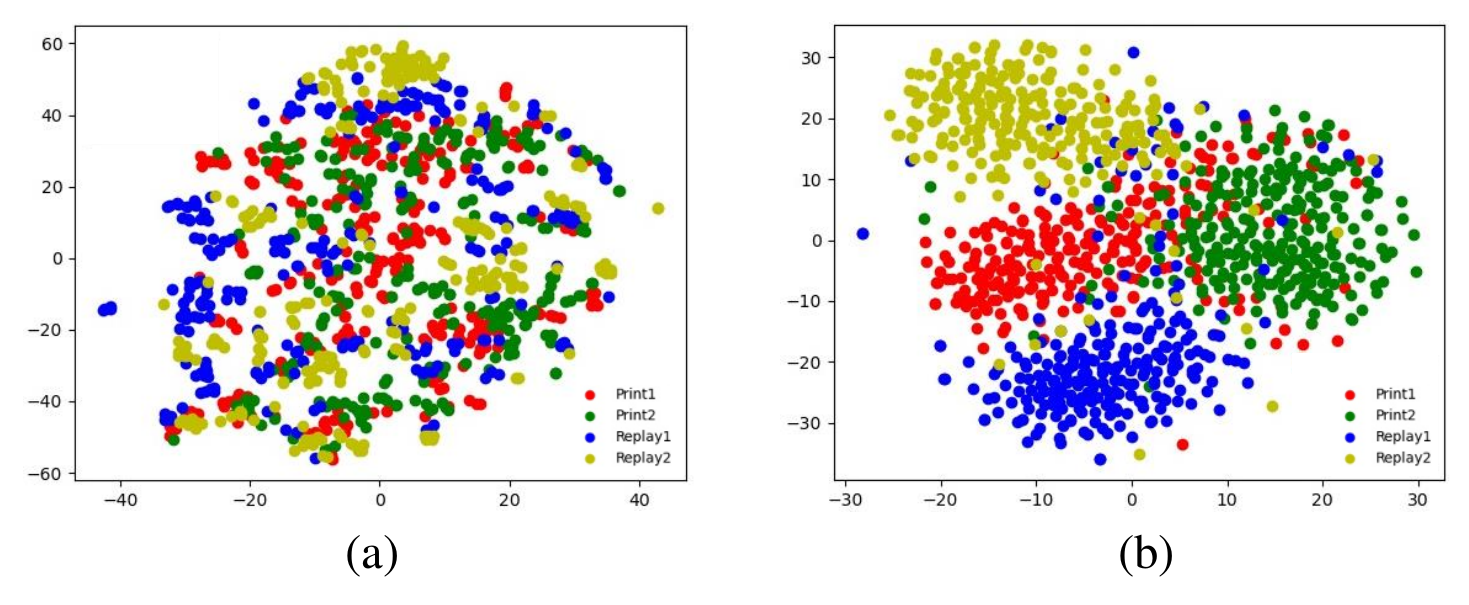}
\caption{
Visualization of the spoofing pattern representations $\hat{z}^{t}_{s}$ in the latent space. (a) is the result of DSDG without the classifier and the orthogonal loss $\mathcal{L}_{\text {ort}}$, while (b) is the result of well-trained DSDG.
}
\label{spoof visualtion}
\vspace{-0.5em}
\end{figure}
\textbf{Cross-dataset Testing.}
In this experiment, CASIA-MFSD and Replay-Attack are used for cross-dataset testing.
We first perform the training on CASIA-MFSD and test on Replay-Attack. As shown in Tab.~\ref{CASIA-Replay}, our method achieves competitive results and is better than CDCN. Then, we switch the datasets for the other trial, and obtain a significant improvement (of 5.9\% HTER) compared with CDCN. Note that, FAS-SGTD performs anti-spoofing on video-level, and the result of our method (26.7\% HTER) is the best among all frame-level methods.
\subsection{Analysis and Visualization}
\label{visual}
\textbf{Visualization of Generated Images.}
\label{gen_vis}
We visualize some generated images on the OULU-NPU Protocol 1 by DSDG in Fig.~\ref{render}(b). The generated paired images provide the identity diversity that the original data lacks. Moreover, DSDG successfully disentangles the spoofing patterns from the real spoofing images and preserves them in the generated spoofing images. We also provide the generated results on SiW-M, which are shown in Fig.~\ref{render-siw}. SiW-M contains more diverse presentation attacks (\textit{e.g.} Replay, Print, 3D Mask, Makeup Attacks and Partial Attacks) and less identities in some presentation attacks. Even so, DSDG still generates the spoofing images with diverse identities, which retain the original spoofing patterns.
\par
\begin{figure}[t]
\centering
\includegraphics[scale=0.72]{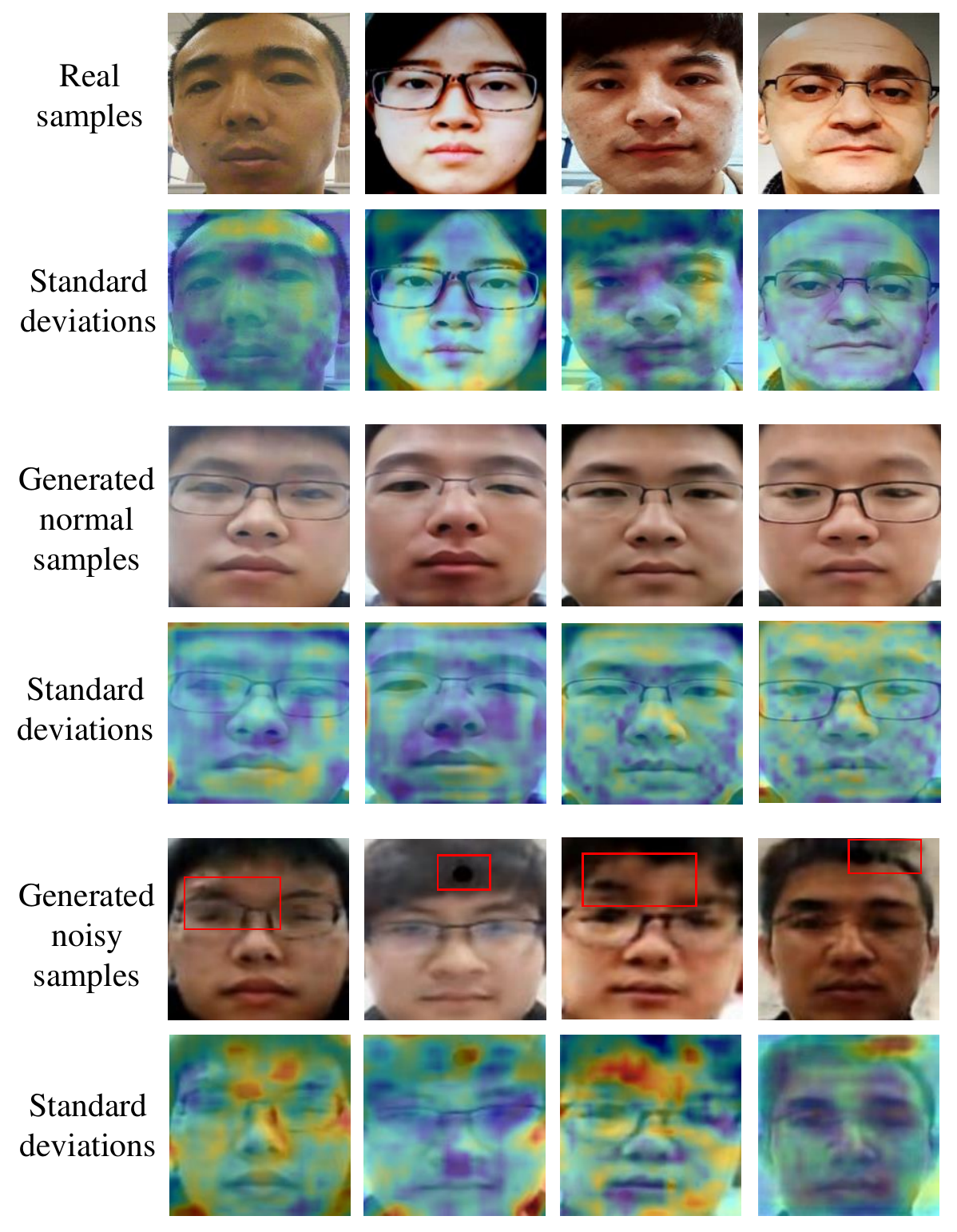}
\caption{Visualization of the standard deviation predicted by DUM. The 1st, 3rd and 5th rows are real samples, generated normal samples and generated noisy samples, respectively. The 2rd, 4th and 6th are samples blended with the standard deviations.
}
\label{u_heatmap}
\vspace{-0.5em}
\end{figure}
\textbf{Visualization of Disentanglement on OULU-NPU.}
To better present the disentanglement ability of DSDG, we fix the identity representation, sample diverse spoofing pattern representations, and generate the corresponding images.
Some generated results are presented in Fig.\ref{render-oulu}, where the odd rows are live images with the same identity, and the even rows contain spoofing images with various spoof types. Obviously, DSDG disentangles the spoofing pattern and the identity representations. Furthermore, we used t-SNE to visualize the spoofing pattern representations ${z}^{t}_{s}$ by feeding the test spoofing images to the $Enc_{s}$. Firstly, we discard the classifier and the orthogonal loss $\mathcal{L}_{\text {ort}}$ when training the generator, and present the distributions of spoofing pattern representations in Fig.~\ref{spoof visualtion}(a). We can observe that the distributions are mixed together. Then, the same distributions of well-trained DSDG are shown in Fig.~\ref{spoof visualtion}(b). Obviously, the distributions of DSDG are well-clustered to four clusters correspond to four spoof types. 
\par
\textbf{Visualization of Standard Deviation.}
\label{heatmap}
Fig.~\ref{u_heatmap} shows the visualization of the standard deviations predicted by DUM, where the red area indicates high standard deviation. The 1st and 3rd rows are images with good quality whose standard deviations are relatively consistent. The 5th row contains some noisy samples with distorted regions indicated in red boxes. Obviously, the distorted regions have higher standard deviations, as shown in the 6th row. Hence, DUM provides the ability to estimate the reliability of the generated images.
\par
\begin{figure}[t]
\centering
\includegraphics[scale=0.43]{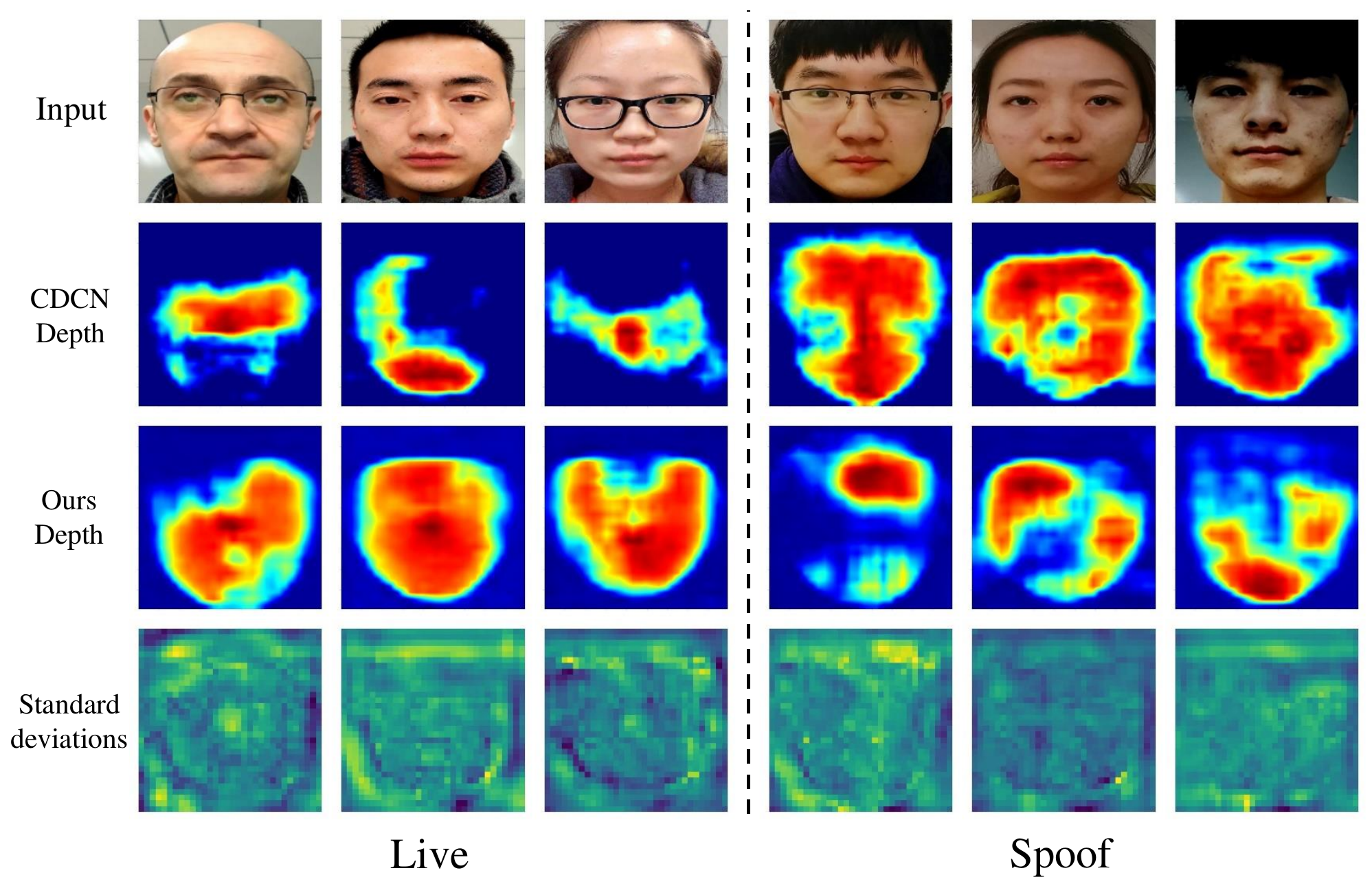}
\caption{Some examples that are failed by CDCN~\cite{Yu2020SearchingCD}, but successfully detected with our method. The rows from top to bottom are the input images, the predicted depth maps by CDCN, the predicted depth maps by our methods and the predicted standard
deviations by CDCN with DUM, respectively.}
\label{compare}
\vspace{-1em}
\end{figure}

\textbf{Visualization Comparison with CDCN~\cite{Yu2020SearchingCD}}
In Fig.~\ref{compare}, we present some hard samples on  OULU-NPU P4, which focuses on the evaluation across unseen environmental conditions, attacks and input sensors. It can be seen that some ambiguous samples are difficult for CDCN, but are predicted correctly by our method, further demonstrating the effectiveness and the generalization ability of the proposed method. We also visualize the corresponding standard deviations of the raw CDCN with DUM to explore the reason why DUM can boost the performance of the raw CDCN. It can be observed that some reflective areas have relatively larger standard deviations. Besides, the edge areas of the face and the background also have larger standard deviations. Obviously, these areas are relatively difficult to predict the precise depth. However, the model can estimate the depth uncertainty of these ambiguous areas with the DUM and predict more robust results. That is the reason why only adopt DUM can also improve the performance of the model.
\par
\section{Conclusions}
Considering that existing FAS datasets are insufficient in subject numbers and variances, which limits the generalization abilities of FAS models, in this paper, we propose a novel Dual Spoof Disentanglement Generation (DSDG) framework that contains a VAE-based generator. DSDG can learn a joint distribution of the identity representation and the spoofing patterns in the latent space, thus is able to preserve the spoofing-specific patterns in the generated spoofing images and guarantee the identity consistency of the generated paired images. With the help of DSDG, large-scale diverse paired live and spoofing images can be generated from random noise without external data acquisition. The generated images retain the original spoofing patterns, but contain new identities that do not exist in the real data. We utilize the generated image set to enrich the diversity of the training set, and further promote the training of FAS models. However, due to the defect of VAE, a portion of generated images have partial distortions, which are difficult to predict precise depth values, degenerating the widely used depth supervised optimization. Thus, we introduce the Depth Uncertainty Learning (DUL) framework, and design the Depth Uncertainty Module (DUM) to alleviate the adverse effects of noisy samples by estimating the reliability of the predicted depth map. It is worth mentioning that DUM is a lightweight module that can be flexibly integrated with any depth supervised training. Finally, we carry out extensive experiments and adequate comparisons with state-of-the-art FAS methods on multiple datasets. The results demonstrate the effectiveness and the universality of our method. Besides, we also present abundant analyses and visualizations, showing the outstanding generation ability of DSDG and the effectiveness of DUM.

\section*{Acknowledgement}
This work was supported by the National Key R\&D Program of China under Grant No.2020AAA0103800.
\ifCLASSOPTIONcaptionsoff
  \newpage
\fi



%

\bibliographystyle{IEEEtran}
\bibliography{egbib}


%










\end{document}